\documentclass[lettersize,journal]{IEEEtran}

\usepackage[caption=false,font=normalsize,labelfont=sf,textfont=sf]{subfig}
\usepackage{graphicx}
\usepackage{amsmath,amsfonts}
\usepackage{algorithmic}
\usepackage{algorithm}
\usepackage{array}
\usepackage{textcomp}
\usepackage{url}
\usepackage{verbatim}
\usepackage{cite}
\usepackage{multirow}
\usepackage{amssymb}
\usepackage{booktabs}
\usepackage{color}
\usepackage{ragged2e}
\usepackage{hyperref}
\usepackage{bbding}
\usepackage{adjust box}
\usepackage{stfloats}
\usepackage{rotating}
\begin{document}

\title{ObjectAdd: Adding Objects into Image via a Training-Free Diffusion Modification Fashion}

\author{Ziyue Zhang,
        Mingbao Lin,
        Quanjian Song,
        Yuxin Zhang,
        Rongrong Ji,~\IEEEmembership{Senior Member,~IEEE}
        \thanks{Manuscript received Feb. XX, 2025.  (Corresponding author: Rongrong Ji)}

\iffalse
\IEEEcompsocitemizethanks{

\IEEEcompsocthanksitem Z. Zhang, Q. Song and Y. Zhang is with the Key Laboratory of Multimedia Trusted Perception and Efficient Computing, Ministry of Education of China, Xiamen University, China.\protect
\IEEEcompsocthanksitem M. Lin is with the Skywork AI, Singapore.\protect
\IEEEcompsocthanksitem R. Ji (Corresponding  Author) is with the Key Laboratory of Multimedia Trusted Perception and Efficient Computing, Ministry of Education of China, Xiamen University, China, also with Institute of Artificial Intelligence, Xiamen University, Xiamen 361005, China (e-mail: rrji@xmu.edu.cn).\protect
}
\fi
}

\markboth{}%
{}

\maketitle

\begin{abstract}
We introduce ObjectAdd, a training-free diffusion modification method to add user-expected objects into user-specified area.
The motive of ObjectAdd stems from: first, describing everything in one prompt can be difficult, and second, users often need to add objects into the generated image. 
To accommodate with real world, our ObjectAdd maintains accurate image consistency after adding objects with technical innovations in:
(1) embedding-level concatenation to ensure correct text embedding coalesce;
(2) object-driven layout control with latent and attention injection to ensure objects accessing user-specified area;
(3) prompted image inpainting in an attention refocusing \& object expansion fashion to ensure rest of the image stays the same.
With a text-prompted image, our ObjectAdd allows users to specify a box and an object, and achieves: (1) adding object inside the box area; (2) exact content outside the box area; (3) flawless fusion between the two areas. The project is at \url{https://github.com/potato-kitty/ObjectAdd}.
\end{abstract}


\section{Introduction}
\IEEEPARstart{R}{ecent} years have seen significant advancements in transformer architecture~\cite{vaswani2017attention}, like CLIP~\cite{radford2021learning}, which have improved natural language embedding methods. Additionally, the development of diffusion models~\cite{ho2020denoising} has led to more accurate and diverse image generation. These advancements have jointly made text-to-image generation increasingly popular. Typically, a text-to-image model takes a description sentence as input and generates a description-aware image. Models such as DALL-E 3~\cite{betker2023improving} and Stable Diffusion~\cite{rombach2022high} have been particularly noteworthy for their impressive performances.
On the premise of these developments, many software platforms have been built to enable non-professionals to easily generate images from text, which has a significant impact on the art industry.

\begin{figure}[!t]
  \centering
  \subfloat[]{\includegraphics[height=7cm]{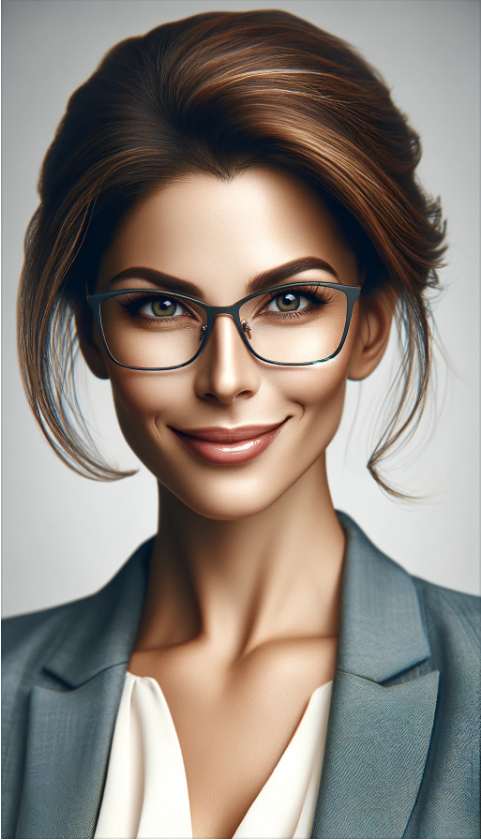}}
  \label{fig:DALL-E 3-a}
  %\end{subfigure}
  \hfill
  \subfloat[]{\includegraphics[height=7cm]{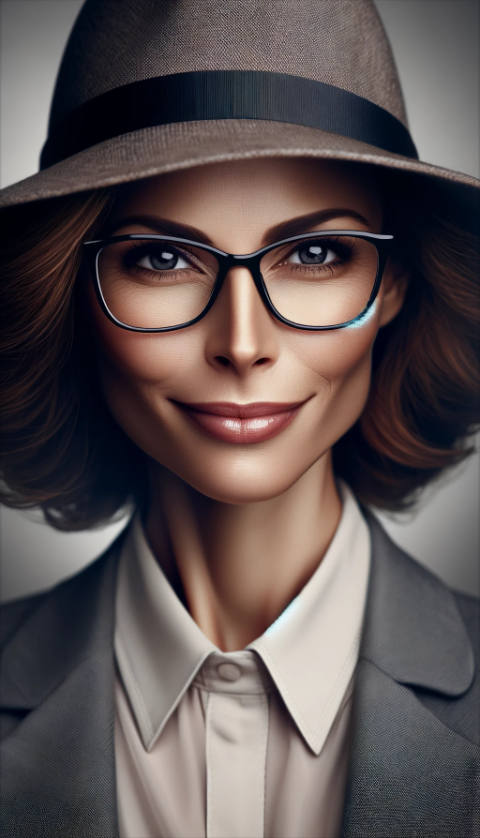}}
  \label{fig:DALL-E 3-b}
  %\end{subfigure}
  \caption{Image generation of ChatGPT-4. Image (a) is prompted by ``A woman wearing glasses.'' while image (b) generates from subsequent ``Now let her wear a hat.'' Despite the high-qualified image (a), adding a hat causes visual inconsistency in image (b).}
  \label{fig:DALL-E 3}
  \vspace{-0.5em}
\end{figure}

One example is Midjourney, where users can create and share AI-generated images~\cite{midjourney}. Another example is ChatGPT, of which the latest version allows users to request image generation~\cite{chatGPT}. Even classical software tools like Photoshop have improved their functionality by incorporating AI technology, enabling them to perform previously unimaginable tasks, such as changing the orientation of objects in photos~\cite{photoshop}. These applications pave the way for the future of artificial general intelligence (AGI) as they can be operated solely through natural language commands. With the use of these tools, even individuals without any formal training can now create their own artwork, leading to a significant boost in the efficiency of the art industry.
However, most text-to-image models rely solely on text prompts as inputs. This makes it challenging for users to convey spatial relationships between objects. Take the most famous large model, ChatGPT, for example. It requires users to describe all their requirements through chatting, which is not an easy task for non-professional personnel.
Some approaches have been proposed to solve this problem, with ControlNet~\cite{zhang2023adding} possibly being one of the most famous models. ControlNet allows users to include canny edge or human pose as extra inputs, enabling control over the pose of humans or animals in the generated image. With the use of ControlNet, users can control the generated result in a more intuitive way, but it still requires users to decide the layout all at once before seeing any outcome result.

Thus, we conclude two practical issues:
first, describing everything in a single prompt can be difficult, and second, users often need to modify the generated image multiple times to achieve their desired results. Unfortunately, current models are stuck in these challenges. For example, as shown in Fig.\,\ref{fig:DALL-E 3}, in our tests with ChatGPT-4, which uses DALL-E 3 for image generation, we find that while the initial images are of high quality, but they lose consistency after modification. Details such as the orientation of a person and their clothing are lost.

In this paper, we propose ObjectAdd, a training-free method that allows users to add an object into a selected area of an image, while maintaining the rest content almost unchanged.
Given an image prompt and object word, we start by analyzing the negative impact of self-attention mechanism in the text encoder, causing embedding confusion if conducting a simple text concatenation. 
Therefore, the first innovation of our ObjectAdd is to attempt concatenating the embedding outputs and amalgamate information from both texts while preventing them from interfering with each other. 
Then, we continue our second innovation of object-driven layout control in order to capture objects within the user-drawn area.
ObjectAdd takes into consideration both latent- and attention-level information into the diffusion process of the edited image.
For latent information injection, an object-oriented diffusion process is individually conducted, during which, the training-free backward guidance~\cite{chen2023trainingfree} is firstly adopted to strengthen the cross-attention score of added objects inside the user-drawn box, then the corresponding area in diffusion latent is transferred to that of edited image.
As for attention information injection, we enhance the cross-attention map in the user-drawn area of edited image.
By doing so in the early diffusion steps, we realize adding a high-quality object to the edited image.

The final innovation of our ObjectAdd lies in prompted image inpainting to reach the  goal that the rest content should be kept unchanged with the original image. This is accomplished by shifting the mask attention from user-drawn box area to the generated object by way of attention refocusing and object expansion. We observe a highly-responsive attention map of the added object in the middle diffusion steps. Therefore, we perform clustering upon the object related attention map and pick up the object-centralized sub-figure that intersects mostly with user-drawn area.
With an approximate location of the object, we locate seed pixels that fall into the object-centralized area but at least one of its neighbors falls out of the object area. We recover the out-of-object neighbors if they are close to its seed pixel.
Applying attention refocusing and object expansion in a single middle diffusion step well attains the out-of-box content.
For user convenience, we also provide an extension of our method that allows real images as the input. To achieve this, we first reshape the input real image to fit the user-selected area and then apply a newly proposed inversion method~{\color{blue}\cite{}} before utilizing our ObjectAdd methods. While all other modifications remain the same as in the case of non-real image inputs, we replace the mask for the user-selected area with the segmentation of the reshaped input real image. Our experiments indicate that this approach results in improved performance.

\section{Related Works}

\subsection{Training-Free Control}
Since large text-to-image models are trained with extensive datasets, they are generally demonstrated to embrace enough information for most tasks. 
Given that retraining a large model can be challenging and may lead to catastrophic forgetting~\cite{zhai2024investigating}, many researchers focus on training-free methods to control the model outputs. These methods fall into two main categories: forward and backward guidance. In our work, we combine both approaches.

\subsubsection{Forward Guidance.} To modify the output of a pretrained model without retraining, a straightforward approach is to directly adjust the parameters without employing gradient descent. Since this approach doesn't involve backward propagation, they are referred to as ``forward guidance.'' 
Most diffusion-based models incorporate attention maps, which contain semantic information~\cite{hertz2022prompt}. 
Many forward guidance methods modify attention maps to improve their effectiveness~\cite{hertz2022prompt,densediffusion,patashnik2023localizing}. The two main methods for editing attention maps are enhancing and swapping. Enhancement is often applied to the cross-attention map. This increases the value in the target area of the map, ensuring that objects are generated correctly~\cite{densediffusion}. Swapping, however, can be applied to both self- and cross-attention maps. In some studies~\cite{hertz2022prompt}, researchers swap parts of the cross-attention map to replace one object with another, while keeping the rest of the image unchanged. In other cases~\cite{patashnik2023localizing}, parts of the self attention map from the original image are swapped with the editing output to maintain consistency.

\subsubsection{Backward Guidance.} 
Although researchers sometimes expect to eliminate training, having a loss function is still an easy way to guide the model. Methods that require no training but utilize a loss function and gradient descent are known as ``backward guidance.'' These methods are commonly used to control the layout of generated images~\cite{chen2023trainingfree,chefer2023attend,xie2023boxdiff}. Compared to forward guidance, backward guidance imposes stricter constraints on object placement. It typically identifies object locations using cross-attention maps~\cite{chen2023trainingfree,chefer2023attend,xie2023boxdiff}, and calculates loss based on user-marked boxes or areas. The loss value is then used to apply gradient descent to the latent, which is updated iteratively at each time step. This process continues until it reaches the maximum number of iterations or the loss falls below a certain threshold. Backward guidance is also effective for repositioning objects. For instance, in DragDiffusion~\cite{shi2023dragdiffusion}, it is used to move selected objects in a specific direction, and user-drawn boxes are therefore no longer necessary.

\subsection{Training-Dependent Image Editing}
Though fine-tuning a large model could be difficult, an external model can be trained to assist the larger model. For instance, in ITI-Gen~\cite{zhang2023iti}, researchers train an encoder to obtain distinctive token embeddings. This is done to accurately generate human faces that match the descriptive words in the prompt, such as skin tone. Another example is seen in GLIGEN~\cite{li2023gligen}, where an extra self-attention layer is trained and integrated into the pretrained model. This layer encodes and transmits location information to the model. The benefit of this method is its stability compared to training-free approaches. However, it requires a dataset tailored to its specific task, which can sometimes be challenging to obtain. 
InstructPix2Pix~\cite{brooks2023instructpix2pix}  could be the one that has a most similar goal to our approach. It allows users to specify their desired modifications to an input image through prompts. This model takes input image as conditional vector, as well as the instruction prompt indicated by user. The output of this approach is the edit version of input image, following the instruction. However, it still only allows prompts as input for specifying the requirements, and its results often fail to guarantee the consistency.

\subsection{Decoupled Text Embedding}\label{bg_txt_embedding}
Most current popular text encoders include self-attention layers, implying that the encoding of a word can be influenced by other words within the input. This poses a challenge known as prompt concatenation. Simply combining two prompts and encoding them together often cause confusion. For instance, consider the two prompts ``a cat'' and ``a dog'' combined to form a new prompt ``a cat a dog.'' This combination lacks in-context meaning, making it challenging for the model to comprehend.
The solution to this problem involves encoding the prompts separately and then concatenating their embeddings. The selection of separately encoded words may vary; in the case of ITI-GEN~\cite{zhang2023iti}, category-related words are encoded individually, while in another paper~\cite{feng2022training}, all nones are encoded individually. Both of the papers observe increase performance.

\subsection{Diffusion Based Real Image Editing}\label{real_image_edit}
Diffusion models have demonstrated significant capabilities in controllable image generation~\cite{hertz2022prompt,cao_2023_masactrl}. There is a growing interest in using them for real image editing. Since diffusion models start with noise as input, the primary challenge to edit real images is to convert the input real image into noise in such a way that the noise can be reconstructed into the original image after the denoising steps. This conversion process is called inversion. The most commonly used method today is DDIM inversion~\cite{couairon2023diffedit}, which inverts each denoising step by adding the predicted noise to the latent representation. Several approaches have been proposed to improve the quality of inversion. In~\cite{pan2023effective}, the researchers suggested that inversion could be considered a fixed-point iteration problem and could be solved using iterative updating methods. ReNoise~\cite{garibi2024renoise} addresses the inversion problem from a similar perspective, but introduces minor biases in each step to enhance the edit-ability of the inversion results.

Once the real image has been inverted to noise, it can be used as the input for the diffusion model. Various methods can then be applied to control the generation process, thereby editing the real image. MasaCtrl \cite{cao_2023_masactrl} is one example. In that paper, researchers replace information in the editing branch with the corresponding information in the reconstruction branch to maintain consistency while using new prompts as conditions for generation. This approach allows editing original image, applying modifications while retaining most of the original information. %An example provided in MasaCtrl is transforming an image of a sitting dog into an image of a running dog. %That indicate the weakness of this method, it can only apply minor modifications, while been incapable for major changes like insert an object as what we achieve in this paper.

\section{Methodology}\label{sec:method}
\subsection{Preliminaries}\label{Pre}
Given a user-specified prompt $\mathcal{P}$, a stable diffusion model $F$ performs a series of $T$ denoising steps upon a Gaussian noise $\mathcal{I}^{T} \in \mathbb{R}^{H \times W \times C}$, which finally evolves into a customized image $\mathcal{I}^0 \in \mathbb{R}^{H \times W \times C}$ conditioned on $\mathcal{P}$:
\begin{equation}\label{eq:imgp}
\mathcal{I}^0 = F(\mathcal{I}^T;\mathcal{P}), \qquad \mathcal{I}^{T} \sim \mathcal{N}(0, \mathbf{I}). 
\end{equation}

Herein, $H$, $W$ and $C$ respectively indicate image height, width and channel number.
Inside the stable diffusion model $F$, there are some intermediate representations that are the focus in our method:
(1) Text embedding $\mathcal{E}_{\mathcal{P}} \in \mathbb{R}^{N \times D}$, usually derived from a CLIP model~\cite{radford2021learning}, where $N$ represents the maximum token sequence length in the text encoder and $D$ denotes the embedding dimension;
(2) The $t$-th step latent denoised representation $\mathcal{I}^t \in \mathbb{R}^{H \times W \times C}$;
%
%(3) The $\gamma$-layer self-attention map at the $t$-th step $\mathcal{S}^{\gamma t} \in \mathbb{R}^{(H_{\gamma} \cdot W_{\gamma}) \times (H_{\gamma} \cdot W_{\gamma})}$ where $H_{\gamma}$ and $W_{\gamma}$ stand for image height and width in the $\gamma$-th layer;
%
(3) The $\gamma$-layer cross-attention map at the $t$-th step $\mathcal{C}^{\gamma t} \in \mathbb{R}^{N \times (H_{\gamma} \cdot W_{\gamma})}$, with $\mathcal{C}^{\gamma t}_{i,:} \in \mathbb{R}^{H_{\gamma}\cdot W_{\gamma}}$ indicating the attentive score of the $\gamma$-th layer output to the $i$-th token. For ease of the following representation, we reshape $C^{{\gamma}t}_{i, :}$ into $H_{\gamma} \times W_{\gamma}$.
Notice these symbols will be revised accordingly in the following sections, in order to adapt to outputs of different prompts. For example, $\mathcal{C}^{{\gamma}t}$ will be revised as $\bar{\mathcal{C}}^{{\gamma}t}$ for adapting to Eq.\,(\ref{eq:imgpw}) and $\tilde{\mathcal{C}}^{{\gamma}t}$ for adapting to Eq.\,(\ref{eq:imgw}).

Our intend to edit on image $\mathcal{I}^0$ to derive a user-expected version $\bar{\mathcal{I}}^0$ under two application-oriented settings:
1) a user-drawn box in image $\mathcal{I}^0$ and 2) a user-specified word-level prompt $\mathcal{W}$ describing an object.
The box can be characterised as a binary mask $\mathcal{M} \in \{0, 1\}^{H \times W}$ where $\mathcal{M}_{i, j} = 1$ if pixel $\mathcal{I}^0_{i, j}$ falls into the drawn box, and $\mathcal{M}_{i, j} = 0$ else.
Note that, the $\mathcal{M}$ is downsampled into $\mathcal{M}^{\gamma} \in \mathbb{R}^{H_{\gamma} \times W_{\gamma}}$, for adapting to the $\gamma$-th layer image shape.

The current edited image $\bar{\mathcal{I}}^0$ is featured with: 1) the word $\mathcal{W}$ described object inside the box area; 2) exact content with $\mathcal{I}^0$ outside the box area; 3) flawless fusion between the two areas. The evolution process of $\bar{\mathcal{I}}^0$ can be described as: 
\begin{equation}\label{eq:imgpw}
\bar{\mathcal{I}}^0 = F(\mathcal{I}^T;\{\mathcal{P}, \mathcal{W}\}), \qquad \mathcal{I}^{T} \sim \mathcal{N}(0, \mathbf{I}). 
\end{equation}

To this end, we conclude three pivotal ingredients for the success of $\mathcal{I}^0$, which will be respectively elaborated in the following subsections:
1) correct text embedding coalesce of prompts $\mathcal{P}$ and $\mathcal{W}$;
2) object-driven layout control for $\mathcal{W}$ described objection generation within the user-drawn box area;
3) prompted image inpainting to recover out-of-box area of $\mathcal{P}$ prompted content.

\subsection{Text Embedding Coalesce}
We first deal with the coalesce of prompts $\mathcal{P}$ and $\mathcal{W}$ in Eq.\,(\ref{eq:imgpw}).
As specified in Sec.\,\ref{bg_txt_embedding}, the self-attention mechanism in text encoder causes embedding confusion if simply concatenating two prompts together $concat(\mathcal{P}, \mathcal{W})$.  
Taking $\mathcal{P}$ = ``{\color{blue}A woman wearing glasses}'' and $\mathcal{W}$ = ``{\color{cyan}A hat}'' as a simple example, the concatenation results in prompt ``{\color{blue}A woman wearing glasses} {\color{cyan}A hat}'' tokenized as (``[CLS],'' ``{\color{blue}a},'' ``{\color{blue}woman},'' ``{\color{blue}wearing},'' ``{\color{blue}glasses},'' ``{\color{cyan}a},'' ``{\color{cyan}hat},'' ``[SEP]'') where ``[CLS]'' and ``[SEP]'' respectively denote the start and end tokens. 
For self-attention without causal masks, these tokens are mutually attentioned therefore causing different embeddings of ``{\color{blue}a},'' ``{\color{blue}woman},'' ``{\color{blue}wearing},'' ``{\color{blue}glasses},'' ``{\color{cyan}a},'' and ``{\color{cyan}hat},'' compared to individually embed prompts $\mathcal{P}$ (``[CLS],'' ``{\color{blue}a},'' ``{\color{blue}woman},'' ``{\color{blue}wearing},'' ``{\color{blue}glasses},'' ``[SEP]'') and $\mathcal{W}$ (``[CLS],'' ``{\color{cyan}a},'' ``{\color{cyan}hat},'' ``[SEP]''). Similarly, masked self-attention causes different embeddings of ``{\color{cyan}a}'' and ``{\color{cyan}hat}.''
%
%
%In both cases, diffusion model generates an imperfect hat figure, while the first case causes a different dog image.
%
%
Most current diffusion models deploy a masking-free CLIP-style text encoder~\cite{radford2021learning}, which contradicts the intention in this paper that users add an object into a selected area of an image, while maintaining the rest content unchanged.

Instead of operating the inputs of the text encoder, to reach correct text embedding coalesce, we modify outputs of the text encoder, \emph{i.e.}, text embeddings.
We embed prompt $\mathcal{P}$ and prompt $\mathcal{W}$ separately, as $\mathcal{E}_{\mathcal{P}} \in \mathbb{R}^{N \times D}$ and $\mathcal{E}_{\mathcal{W}}  \in \mathbb{R}^{N \times D}$. %Besides, we also embedding a empty prompt $\mathcal{N}$ to generate $\mathcal{E}_{\mathcal{N}} \in \mathbb{R}^{N \times D}$, in another word, $\mathcal{N}$ contained no conditional information. 
Recall $N$ (usually $N = 77$ in CLIP text encoder) denotes the maximum token sequence length in the text encoder.
Denoting the actual token (no ``[PAD]'', ``[CLS]'' and ``[EOS]'') number of prompt $\mathcal{P}$ as $N_{\mathcal{P}}$, and $N_{\mathcal{W}}$ for prompt $\mathcal{W}$, 
we formulate the text embedding coalesce as:
\begin{equation}
    \mathcal{E}_{\{\mathcal{P}, \mathcal{W}\}} = 
    [(\mathcal{E}_{\mathcal{P}})_{1:N_{\mathcal{P}}+1,:};
    (\mathcal{E}_{\mathcal{W}})_{2:,:}].
\end{equation}

Note that, the $\mathcal{E}_{\mathcal{P}}$ and $\mathcal{E}_{\mathcal{W}}$ start with the same ``[CLS]'' token embedding. 
By concatenating the embedding outputs, we amalgamate information from both prompts while preventing them from interfering with each other. Then, our ObjectAdd method starts the diffusion surgery upon the coalesce $\mathcal{E}_{\{\mathcal{P}, \mathcal{W}\}}$.

\subsection{Object-Driven Layout Control}\label{layout_ctrl}

We continue addressing the second issue of object-driven layout control in Eq.\,(\ref{eq:imgpw}). To generate the word $\mathcal{W}$ described object within the user-drawn box area, our ObjectAdd injects both latent- and attention-level information into the diffusion process of the edited image $\bar{\mathcal{I}}^0$. Below, we denote $k$ as the token sequence id of added object after the text embedding coalesce.

\subsubsection{Latent Information Injection}
To inject latent-level information, we perform an individual denoising process for the word-level prompt $\mathcal{W}$:
\begin{equation}\label{eq:imgw}
\tilde{\mathcal{I}}^0 = F(\mathcal{I}_T;\mathcal{W}), \qquad \mathcal{I}^{T} \sim \mathcal{N}(0, \mathbf{I}).
\end{equation}

To control $\mathcal{W}$ described object accessing the user-drawn box area, we choose the training-free backward guidance method~\cite{chen2023trainingfree}, basic principle of which is to restrict the size and location of the object.
Recall the cross-attention map $\tilde{C}^{{\gamma}t}_{k, :} \in \mathbb{R}^{H_{\gamma} \times W_{\gamma}}$ represents attentiveness of the added object to the $\gamma$-th layer image.
The learning objective can be generally formulated as:
\begin{equation}\label{eq:objective}
    E\left(\tilde{\mathcal{C}}^{{\gamma}t}, \mathcal{M}^{\gamma}, k\right)=\left(1-\frac{\sum_{h=1}^{H_{\gamma}}\sum_{w=1}^{W_{\gamma}} \mathcal{M}_{h, w}^{\gamma} \cdot  \tilde{\mathcal{C}}^{\gamma t}_{k, h, w}}{\sum_{h=1}^{H_{\gamma}}\sum_{w=1}^{W_{\gamma}} \tilde{\mathcal{C}}^{\gamma t}_{k, h, w}}\right)^2.
\end{equation}

Further, the backward propagation process is conducted to guide the latent denoised representation $\tilde{\mathcal{I}}^t$ in Eq.\,(\ref{eq:imgw}) as:
\begin{equation}
    \tilde{\mathcal{I}}^t \leftarrow \tilde{\mathcal{I}}^t - \eta \nabla_{\tilde{\mathcal{I}}^t} \sum_{\gamma \in \Gamma} E\left(\tilde{\mathcal{C}}^{{\gamma}t}, \mathcal{M}^{\gamma}, k\right),
\end{equation}
which iterates several times. The $\Gamma$ represents all chosen layers of the cross-attention maps and $\eta$ is a learning rate.
By shifting the cross-attention focus, it has been well demonstrated~\cite{chen2023trainingfree} that the word $\mathcal{W}$ described object can be mostly confined to the user-drawn box area of latent $\tilde{\mathcal{I}}^t$.

The goal of this paper is to add an object into image $\bar{\mathcal{I}}^0$, prompted by text embedding coalesce $\mathcal{E}_{\mathcal{P}, \mathcal{W}}$. 
Thus, we further inject the above updated latent $\tilde{\mathcal{I}}^t$ into the intended latent $\bar{\mathcal{I}}^t$ at the $t$-th time step as:
%
%
%\[latent_2 = (1 - mask)*lantent_2 + mask*latent_3\]
\begin{equation}\label{eq:latent-injection}
    \bar{\mathcal{I}}^t = (1 - \mathcal{M}) \odot \bar{\mathcal{I}}^t + \mathcal{M} \odot \tilde{\mathcal{I}}^t,
\end{equation}
where $\odot$ denotes the element-wise matrix product.

\begin{figure}[!t]
  \centering
  \begin{minipage}[t]{0.32\linewidth}
    \centering
    \subfloat[]{\includegraphics[height=2.8cm]{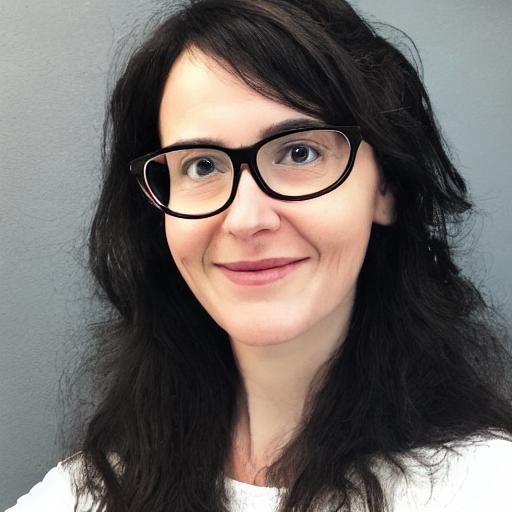}\label{fig:original}}
  \end{minipage}
  \hfill
  \begin{minipage}[t]{0.32\linewidth}
    \centering
    \subfloat[]{\includegraphics[height=2.8cm]{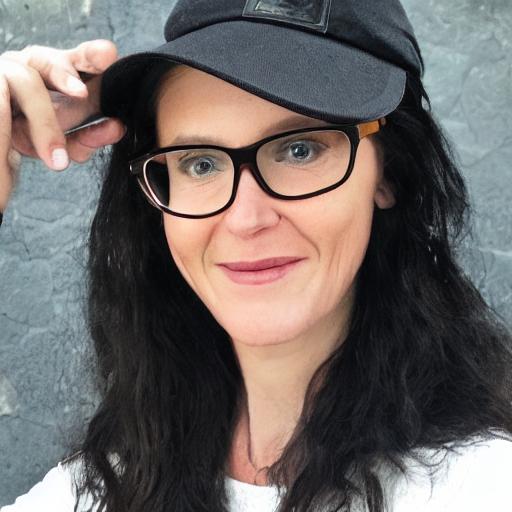}\label{fig:without_object_detection}}
  \end{minipage}
  \hfill
  \begin{minipage}[t]{0.32\linewidth}
    \centering
    \subfloat[]{\includegraphics[height=2.8cm]{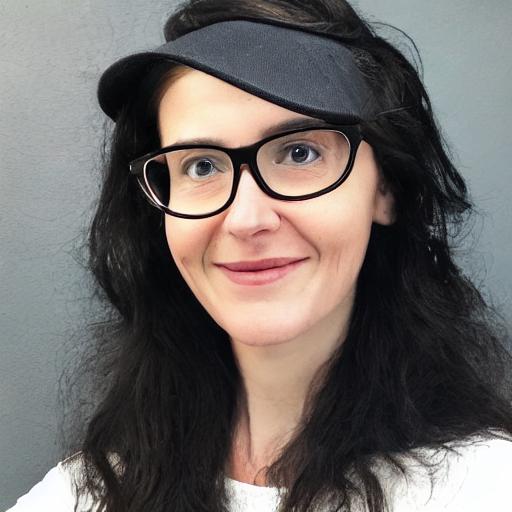}\label{fig:with_object_detection}}
  \end{minipage}
  \caption{Generation comparison under settings of $\mathcal{P}$ = ``A woman wearing glasses'' and $\mathcal{W}$ = ``A hat'':
  (a) $\mathcal{P}$ prompted image $\mathcal{I}^0$; 
  (b) the coalesce $\mathcal{E}_{\{\mathcal{P}, \mathcal{W}\}}$ prompted image $\bar{\mathcal{I}}^0$ after object-driven layout control;
  (c) $\bar{\mathcal{I}}^0$ with prompted image inpainting.
  }
  \label{fig:object_detection}
%  \vspace{-1.0em}
\end{figure}

\subsubsection{Attention Information Injection}
The latent information is injected into edited image $\bar{\mathcal{I}}^0$ by resorting to the third-party information of $\mathcal{\tilde{I}}^0$.
Constructively, our ObjectAdd also injects attention information directly upon image $\bar{\mathcal{I}}^0$ by enhancing the user-drawn area of cross-attention map $\bar{C}^{{\gamma}t}_{k, :} \in \mathbb{R}^{H_{\gamma} \times W_{\gamma}}$.
%
%
%At the $t$-th diffusion timestamp, 
We use the resized mask $\mathcal{M}^{\gamma} \in \mathbb{R}^{H_{\gamma} \times W_{\gamma}}$ to multiply with the average value of $\bar{\mathcal{C}}_{k,:}^{\gamma t}$. The formula of this operation is as following:

\begin{equation}
\bar{\mathcal{C}}_{k,:}^{\gamma t} = softmax\Big(max\big(avg(\bar{\mathcal{C}}_{k,:}^{\gamma t}), 1\big) \cdot \mathcal{M}^{\gamma}\Big).
\end{equation}

Herein, we use a $max$ function to ensure that the result before softmax would not be less than 1 (mask value) for object generation in user-specified area.

We apply information injection across all layers, but only in the first few time steps, to avoid inconsistent style, abrupt edge, \emph{etc}. 
Specifically, as $t = T \rightarrow 0$, latent information injection is applied in the first 20\% time steps while attention information injection is applied in the first 30\% time steps.

\begin{figure}[!t]
  \centering
  \begin{minipage}[t]{0.32\linewidth}
    \centering
    \subfloat[]{\includegraphics[height=70pt]{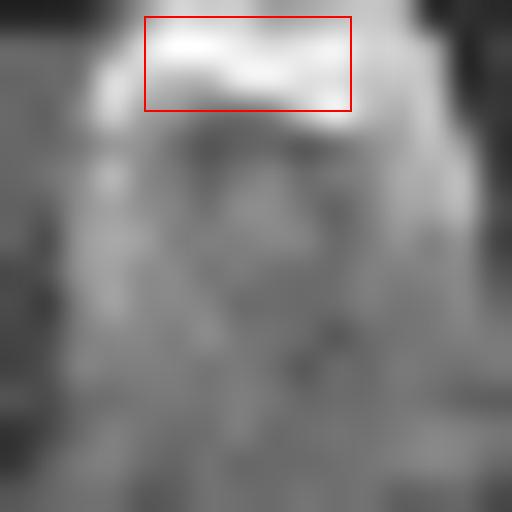}\label{fig:woman_cross}}
  \end{minipage}
  \hfill
  \begin{minipage}[t]{0.32\linewidth}
    \centering
    \subfloat[]{\includegraphics[height=70pt]{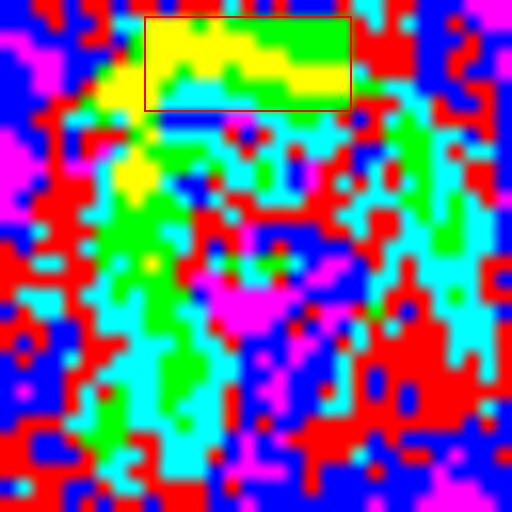}\label{fig:woman_cluster}}
  \end{minipage}
  \hfill
  \begin{minipage}[t]{0.32\linewidth}
    \centering
    \subfloat[]{\includegraphics[height=70pt]{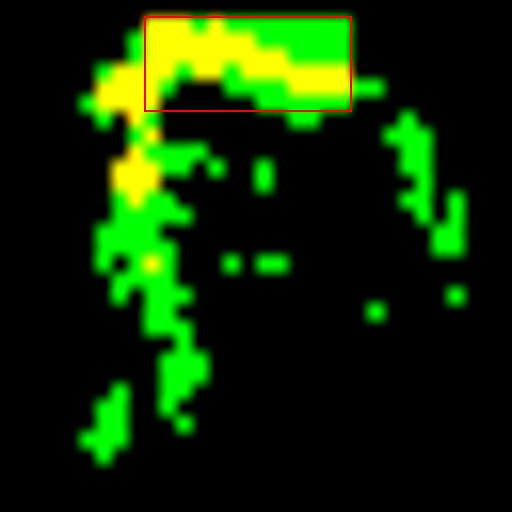}\label{fig:no_ero}}
  \end{minipage}
  \hfill
  \begin{minipage}[t]{0.32\linewidth}
    \centering
    \subfloat[]{\includegraphics[height=70pt]{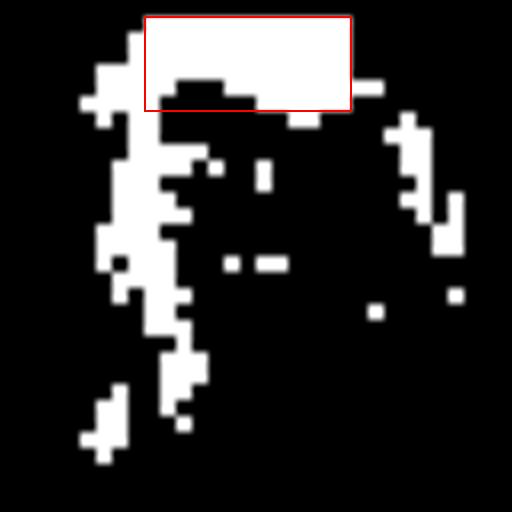}\label{fig:no_ero_msk}}
  \end{minipage}
  \hfill
  \begin{minipage}[t]{0.32\linewidth}
    \centering
    \subfloat[]{\includegraphics[height=70pt]{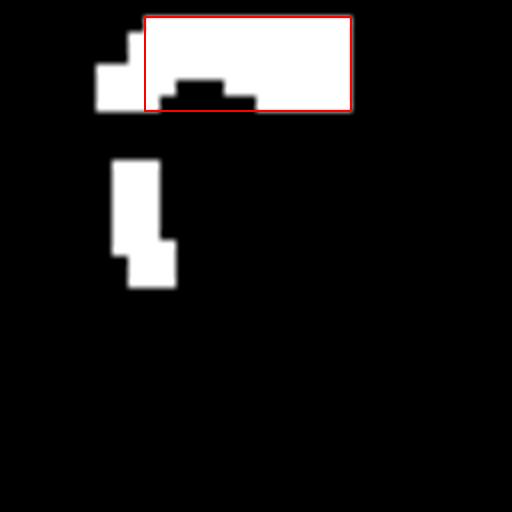}\label{fig:ero_mask}}
  \end{minipage}
  \hfill
  \begin{minipage}[t]{0.32\linewidth}
    \centering
    \subfloat[]{\includegraphics[height=70pt]{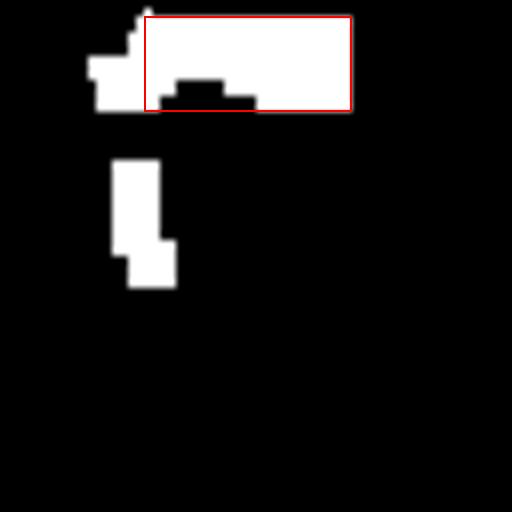}\label{fig:expansion_mask}}
  \end{minipage}
  \caption{Visualization under settings of $\mathcal{P}$ = ``A woman wearing glasses'' and $\mathcal{W}$ = ``A hat''. The user-drawn area is marked by {\color{red}red box}: 
  (a) coalesce $\mathcal{E}_{\{\mathcal{P}, \mathcal{W}\}}$ prompted cross-attention map for object ``hat'';
  (b) clustering of the cross-attention map;
  (c) object-centralized area;
  (d) attention refocusing mask;
  (e) attention refocusing mask with mathematical morphology;
  (f) object expansion mask.
  }
 % \label{fig:cross_attn}
% \vspace{-1.0em}
\end{figure}

\subsection{Prompted Image Inpainting}\label{obj_detect}
Utilizing text embedding coalescence and object-driven layout control, we add a high-quality object to the original image. A prime example is illustrated between Fig.\,\ref{fig:original} and Fig.\,\ref{fig:without_object_detection}, where the prompt $\mathcal{W}$ = ``A hat'' is seamlessly integrated into the user-drawn area. However, Fig.\,\ref{fig:without_object_detection} also serves as a negative example, displaying inconsistencies in background, face, and gesture. To maintain consistency in the out-of-box area, in Fig.\,\ref{fig:with_object_detection}, we employ attention refocusing and object expansion to shift the mask attention from user-drawn box to generated object.

\subsubsection{Attention Refocusing} \label{Attn_refocus}
Our motive mainly originates from an in-depth analysis of the cross-attention map $\bar{\mathcal{C}}_{k, :}^{\gamma t}$ for the added object as visualized in Fig.\,\ref{fig:woman_cross}.
In the middle diffusion steps typically $0.3 \cdot T < t < 0.5 \cdot T$, we find the central area of $\mathcal{W}$ described object will be well focused in the layer of $(H_{\gamma}, W_{\gamma}) = (32, 32)$. %, whatever the prompt $\mathcal{P}$ changes. 
However, the edge area would be mixed up with the out-of-box area.
Therefore, we consider refocusing the attention pixel, to adjust a user-drawn box, \emph{i.e.}, the mask $\mathcal{M}^{\gamma}$, into the object-centralized area $\bar{\mathcal{M}}^{\gamma}$.

To this end, we perform clustering upon the cross-attention map $\bar{\mathcal{C}}_{k, :}^{\gamma t}$ to segment it into $K$ sub-figures, as marked with different colors in Fig.\,\ref{fig:woman_cluster}:
\begin{equation}\label{eq:cluster}
S^1 \cup S^2 \cup ... \cup S^K = \bar{\mathcal{C}}_{k, :}^{\gamma t} \quad \textrm{and} \quad S^1 \cap S^2 \cap ... \cap S^K = \emptyset.
\end{equation}

Then, the object-centralized area consists of sub-figures that: (1) intersecting the most with masked cross-attention map $\mathcal{M}^{\gamma} \odot \bar{\mathcal{C}}_{k,:}^{\gamma t}$:
\begin{equation}
    {argmax}_{S^g} \; | S^{g}  \cap (\mathcal{M}^{\gamma} \odot \bar{\mathcal{C}}_{k,:}^{\gamma t}) |,
\end{equation}
and (2) a large portion falling into the user-drawn area $\mathcal{M}^{\gamma}$:
\begin{equation}\label{eq:portion}
    \frac{|\mathcal{M}^{\gamma} \odot S^i|}{|S^i|} > {\mathcal{H}_1},
\end{equation}
where $\mathcal{H}_1$ is a threshold.
Fig.\,\ref{fig:no_ero} manifests the object-centralized area.
Finally, in Fig.\,\ref{fig:no_ero_msk}, we refocus the user-drawn area by updating the binary mask as $\bar{\mathcal{M}}^{\gamma}$ where $\bar{\mathcal{M}}^{\gamma }_{i,j} = 1$ if attention pixel ($i, j$) falls into object-centralized area and 0 otherwise.
For better performance, in Fig.\,\ref{fig:ero_mask}, we apply mathematical morphology~\cite{soille1999morphological} to remove small objects and fill holes inside the masked area.

\subsubsection{Object Expansion}
Through attention refocusing, we tentatively shift attention from the user-drawn area $\mathcal{M}^{\gamma} \in \mathbb{R}^{H_{\gamma} \times W_{\gamma}}$ to the object-centralized area $\bar{\mathcal{M}}^{\gamma} \in \mathbb{R}^{H_{\gamma} \times W_{\gamma}}$. Unluckily, $\bar{\mathcal{M}}^{\gamma}$ fails to cover all pixels of the object since the cross-attention map helps seize object-centralized pixels but can't well capture the edge area.
For ease of better prompted image inpainting, we upsample $\bar{\mathcal{M}}^{\gamma}$ as  $\bar{\mathcal{M}} \in \mathbb{R}^{H \times W}$ and further propose object expansion in the latent $\bar{\mathcal{I}}^t \in \mathbb{R}^{H \times W \times C}$ space to better model the edge of added objects.

Given $\bar{\mathcal{M}}$ locates the approximate location of the object, we expand the boundary of $\bar{\mathcal{M}}$, as an agent to include the object edge.
For every $\bar{\mathcal{M}}_{i, j}$, we consider its 8-connected component set as $\mathcal{S}_{\bar{\mathcal{M}}_{i, j}} = \{\bar{\mathcal{M}}_{x, y}\}_{x \in \{i, i-1, i+1\}, \, y \in \{j, j-1, j+1\}}$.
Accordingly, we obtain the 8-neighborhood pixels of latent vector $\bar{\mathcal{I}}^t_{i, j} \in \mathbb{R}^{C}$, expressed as $ \mathcal{S}_{\bar{\mathcal{I}}_{i, j}} = \{\bar{\mathcal{I}}_{x, y}\}_{x \in \{i, i-1, i+1\}, \, y \in \{j, j-1, j+1\}}$ where $\bar{\mathcal{I}}^t_{x, y} \in \mathbb{R}^{C}$.
Our object expansion firstly intends to find out the seed pixels that are conditioned on:
\begin{equation}
\begin{split}
    \bar{\mathcal{M}}_{i,j} == 1 \quad \textrm{and} \quad 0 \in \mathcal{S}_{\bar{\mathcal{M}}_{i,j}}, 
\end{split}
\end{equation}
that is, $\bar{\mathcal{I}}_{x,y}^t$ is considered as one seed pixel if it matches the object area and at least one of its neighbors falls out of the object area.
Then, for each neighbor $\bar{\mathcal{I}}_{x, y}$ outside object area with $\bar{\mathcal{M}}_{x, y} == 0$, it indicates a potential edge pixel.
We compute its distance to the seed pixel and the average as:
\begin{equation}
\begin{split}
    D(\bar{\mathcal{I}}_{x, y}| \mathcal{I}_{i,j}, \mathcal{S}_{\mathcal{I}_{i,  j}}) = \\0.5( \cdot dist(\bar{\mathcal{I}}_{x,y}, \; \bar{\mathcal{I}}_{i,j}) + dist\big(\bar{\mathcal{I}}_{x,y}, \; avg(\mathcal{S}_{\mathcal{I}_{i, j}})\big)), \\
   \qquad\; s.t. \quad \bar{\mathcal{M}}_{i,j} == 1, \; \bar{\mathcal{M}}_{x,y} == 0, \\ x \in \{i, i-1, i+1\}, \; y \in \{y, y-1, y+1\},
\end{split}
\end{equation}
where $avg(\cdot)$ averages its inputs and $dist(\cdot, \cdot)$ calculates Euclidean distance. 

We make full use of the distance information to expand the object as:
\begin{equation}
\label{eq:distance}
\bar{\mathcal{M}}_{x, y} = 
\left\{
\begin{array}{ll}
1, \textrm{\; if $D(\bar{\mathcal{I}}_{x, y}| \mathcal{I}_{i,j}$}, \mathcal{S}_{\mathcal{I}_{i,  j}}) < \mathcal{H}_2, \\
\bar{\mathcal{M}}_{x,y}, \textrm{\; otherwise}.
\end{array}\right.
\end{equation}
where $\mathcal{H}_2$ is a user-specified parameter. The mask of out-of-object pixel $\bar{\mathcal{I}}_{x,y}$ is flipped if $\bar{\mathcal{I}}_{x,y}$ is close to the in-object pixel and the average of all neighbors. Every time after going through all seeds $\bar{\mathcal{I}}_{i,j}$, we set those flipped $\bar{\mathcal{I}}_{x,y}$ as new seeds. This process goes iterativly until no $D(\bar{\mathcal{I}}_{x, y}| \mathcal{I}_{i,j}, \mathcal{S}_{\mathcal{I}_{i,  j}}) < \mathcal{H}_2$ found.

As results, we better locate the object pixels, and swap the rest part of the latent to get the optimized latent $\bar{\mathcal{I}}^{t }$, which similar to Eq.\,(\ref{eq:latent-injection}) is formulated as:
\begin{equation}
    \bar{\mathcal{I}}^{t } = (1 - \bar{\mathcal{M}}) \odot {\mathcal{I}}^{t } + \bar{\mathcal{M}} \odot \bar{\mathcal{I}}^{t }.
\end{equation}

%\textcolor{red}{
After this step, the $\bar{\mathcal{I}}^{t}$ is now a combination of detected object and the non-object area of original prompt derived latent $\mathcal{I}^{t}$.
We apply the above swapping at a random time step between $0.3 \cdot T < t < 0.5 \cdot T$.
Since we apply this operation in a time step far away from diffusion end, the robustness of diffusion model would make up the missed detected area like the smaller area in Fig.\,\ref{fig:expansion_mask} and unnatural edges. %caused by this operation.
%}
Details will be ablated in Sec.\,\ref{ablation_exp}.

\begin{figure}[!t]
  \centering
  \begin{minipage}[t]{0.32\linewidth}
    \centering
    \subfloat[]{\includegraphics[height=2.8cm]{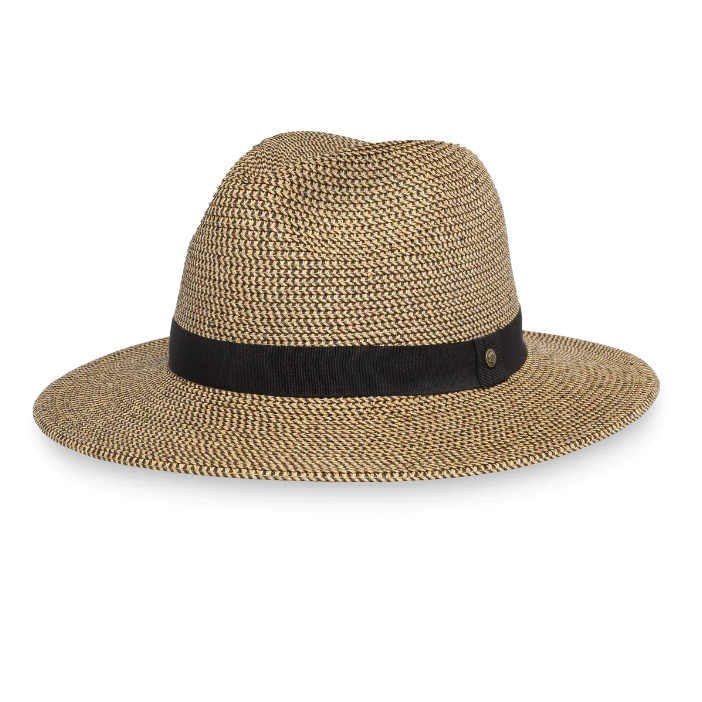}\label{fig:hat_or}}
  \end{minipage}
  \hfill
  \begin{minipage}[t]{0.32\linewidth}
    \centering
    \subfloat[]{\includegraphics[height=2.8cm]{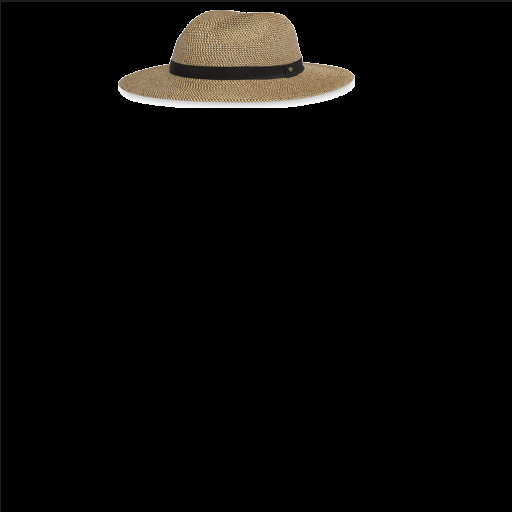}\label{fig:hat_resize}}
  \end{minipage}
  \hfill
  \begin{minipage}[t]{0.32\linewidth}
    \centering
    \subfloat[]{\includegraphics[height=2.8cm]{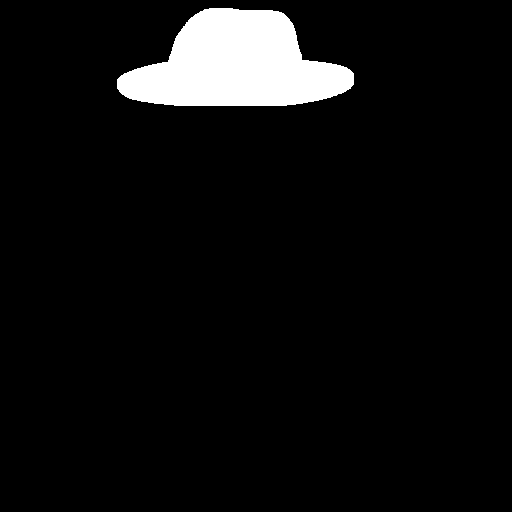}\label{fig:hat_msk}}
  \end{minipage}
  \caption{Pre-processing of input real image.}
  \label{fig:real_img_preprocess}
  \vspace{-1.0em}
\end{figure}

\subsection{Real Image Editing}\label{real_img}

To apply ObjectAdd on real images, we pre-process this extra input, using the results to update the mask and generate the new input in Eq.\,(\ref{eq:imgw}). We omit the layout control methods discussed in Sec.\,\ref{layout_ctrl} since we can directly resize the input.

For the pre-processing of the input real image, as shown in Fig.\,\ref{fig:real_img_preprocess}, we first segment the object and then resize the segmentation to the user-chosen area, as illustrated from Fig.\,\ref{fig:hat_or} to Fig.\,\ref{fig:hat_resize}. We represent the results of this step as $\tilde{\mathcal{Z}}_0$. For convenience, we choose pictures with a white background as examples, allowing for simple segmentation algorithms. For input images with complex backgrounds, a pre-trained segmentation model might be required. We then apply inversion towards $\tilde{\mathcal{Z}}_0$ to obtain the noise $\tilde{\mathcal{Z}}_T$. We use $\tilde{\mathcal{Z}}_T$ to replace the random noise $\mathcal{I}_T$ in Eq.\,(\ref{eq:imgw}), resulting in the following updated equation:
\begin{equation}\label{eq:new_imgw}
\tilde{\mathcal{I}}^0 = F(\tilde{\mathcal{Z}}_T;\mathcal{W}).
\end{equation}

In this way, we ensure that the object in $\tilde{\mathcal{I}}^0$ is generated in the user-specified location after denoising, fulfilling the role of layout control methods discussed in Sec.\,\ref{layout_ctrl}. %Therefore, these methods are not needed in real image editing tasks.

Another modification for real image editing is that, since we have the segmentation of the input, we can generate a more precise mask from it, as shown from Fig.\,\ref{fig:hat_resize} to Fig.\,\ref{fig:hat_msk}. We denote this new mask as $\mathcal{M}'$. Compared to the original mask $\mathcal{M}$, which is represented by the red rectangle area in Fig.\,\ref{fig:woman_cross}, $\mathcal{M}'$ is more accurate. Thus, in real image editing tasks, the mask $\mathcal{M}$ mentioned in previous subsections will be replaced with $\mathcal{M}'$.

For using whatever inversion methods, there are always errors during both inverse and denosing process. For this reason, we inject the corresponding inverted latent $\tilde{\mathcal{Z}}_{\tilde{t}}$ at time-step $\tilde{t}$ where $\tilde{t}$ is set to 39 in this paper, the detail of such injection is as following:
\begin{equation}\label{eq:inverse_inject}
\bar{\mathcal{I}}^{\tilde{t}} = (1 - \mathcal{M}') \odot \bar{\mathcal{I}}^{\tilde{t}} + \mathcal{M}' \odot \tilde{\mathcal{Z}}_{\tilde{t}}.
\end{equation}

%In such way we reduce the error occurred during denoising, although the error caused by inversion would still remain.
%
%Apart from these changes, all other methods remain the same as those introduced for real image editing.

\section{Experimentation}
\subsection{Experimental Setup}\label{setup}
\subsubsection{Implementation Details}
We conduct experiments on one Nvidia GeForce RTX 3090 GPU. In the case of object-driven layout control, as the diffusion step $t = T \rightarrow 0$, we incorporate latent information during the initial 20\% of time steps, while the injection of attention information occurs within the first 30\% of time steps. We employ the strategies of attention refocusing and object expansion to recover the prompted image at the step of $t = 15$, adjustable provided $0.3\cdot T < t < 0.5\cdot T$. Pertaining to the clustering operation in attention refocusing, we establish the cluster number $K = 6$ in Eq.\,(\ref{eq:cluster}) and set the threshold $\mathcal{H}_1 = 0.35$ in Eq.\,(\ref{eq:portion}) along with the parameter $\mathcal{H}_2 = 5$ in Eq.\,(\ref{eq:distance}).
%\footnote{Codes in the {\color{blue}supplementary material} will be publicly released once paper accepted.}

\subsubsection{Compared Methods}
Our ObjectAdd builds upon SD-v1-4~\cite{rombach2022high} as a pre-trained diffusion model, generating outputs at a $512 \times 512$ resolution. As the first attempt to incorporate objects into diffusion-generated images, we compare our method with relevant works: DALL-E 3~\cite{betker2023improving}, P2P~\cite{hertz2022prompt}, and SD-v1-4~\cite{rombach2022high}. Since these methods don't cater to our task, we merge prompts $\mathcal{P}$ and $\mathcal{W}$ into a new prompt $\mathcal{P^{'}}$. For example, if $\mathcal{P} =$ ``A lake'' and $\mathcal{W} =$ ``A boat'', then $\mathcal{P^{'}} =$ ``A lake A boat''. Notably, P2P has a feature that highlights specific tokens, which we utilize to emphasize the second word-level prompt, \emph{i.e.}, $\mathcal{W}$.

\subsubsection{Evaluation Metrics}

Our ObjectAdd requires multi-modal inputs, making extensive dataset evaluation challenging due to annotation demands. As an initial attempt, we assess our approach on a limited input set, comparing it with other methods both qualitatively and quantitatively. For qualitative evaluation, we examine output image intricacies, highlight alterations, and provide subjective assessments of different models' results. Quantitatively, we introduce three metrics to determine average quality and consistency across approaches.

The first one is the ``By Pixels''. We firstly remove the user-drawn area from both the original image, ${\mathcal{I}}^0$, and the edited image, $\bar{\mathcal{I}}^0$. Then, we calculate the average absolute difference between each pixel of two images, formulated as:
\begin{equation}
    \mathcal{L}_{pixels} = avg(\lvert(1 - \mathcal{M}) \odot {\mathcal{I}}^0 - (1 - \mathcal{M}) \odot \bar{\mathcal{I}}^0 \rvert),
\end{equation}
which calculates the consistency of non-edited area. 

Then, in order to assess the quality of the added object, we compute the text-to-image consistency using the CLIP Score as a metric:
\begin{equation}
    \mathcal{L}_{CLIP} = clip(\mathcal{M} \odot \bar{\mathcal{I}}^0,\mathcal{W}),
\end{equation}
where the $clip()$ embeds the image and text into the same space, then calculates their distance. The edited images undergo masking to isolate the object. Our script for computing this metric is derived from the open-source codes~\cite{taited2023CLIPScore}.

We also apply FID~\cite{heusel2017gans} metric, since it is a commonly used one for measuring image generation quality.

\subsubsection{Data Preparation}
To facilitate quantitative comparison, we have created our own dataset. The first step involves generating prompts. We provide a few example prompts such as ``A woman wearing glasses'' and ``A lake'' to ChatGPT and request it to generate 320 similar prompts. Using these prompts, we then generate corresponding images. During this step, we fix the random seeds to ensure reproducibility, allowing us to regenerate the same images given the same prompts. 

Following image generation, we use a custom script to draw bounding boxes on these images. The box information for each image is saved in a text file with five lines, detailing the coordinates of the top-left corner, the width and height of the box, and the prompt used to add the object in the final line. In cases where the original image is of low quality, we discard it, resulting in a final dataset of 229 images.

Our model reads these text files according to the prompts, generates the same original image $\mathcal{I}$, and uses the information in the text file to create the mask $\mathcal{M}$, subsequently generating the reference image $\tilde{\mathcal{I}}$.

\begin{figure*}[!t]
    \centering
    \includegraphics[width=0.98\linewidth]{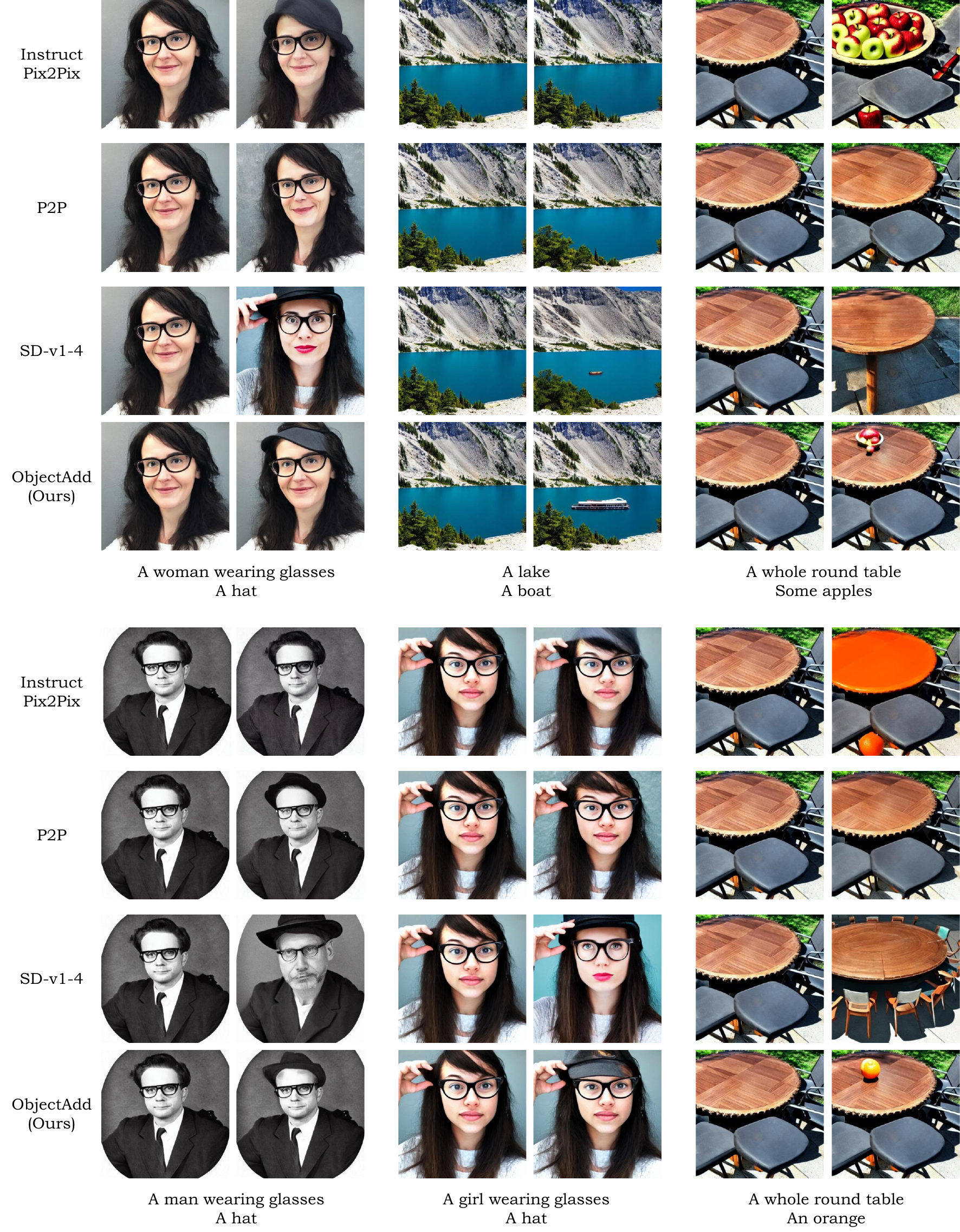}
    \caption{A visual comparison of different methods \emph{w.r.t.} prompt $\mathcal{P}$ generated image and corresponding version of adding word $\mathcal{W}$ described object. Best view with zooming in.}
    \label{result_compare}
    \vspace{-1.0em}
\end{figure*}

\begin{figure*}[!t]
%\captionsetup[subfigure]{labelformat=empty}
  \centering
  \subfloat[\scriptsize{A man wearing glasses \\ A hat}]
  {\includegraphics[width=0.25\textwidth]{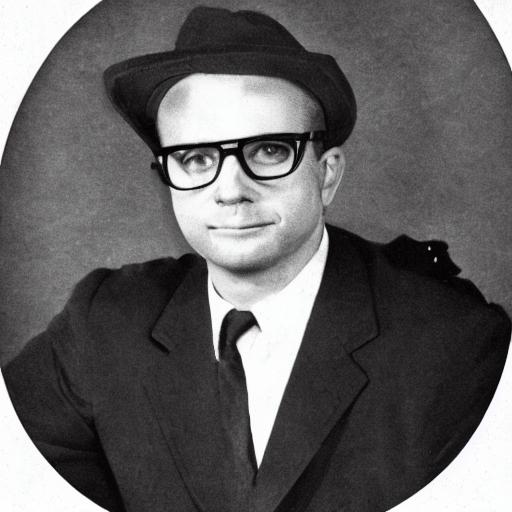}}
  \hspace{0.6cm}
  \subfloat[\scriptsize{A young woman wearing glasses \\ A beret}]
  {\includegraphics[width=0.25\textwidth]{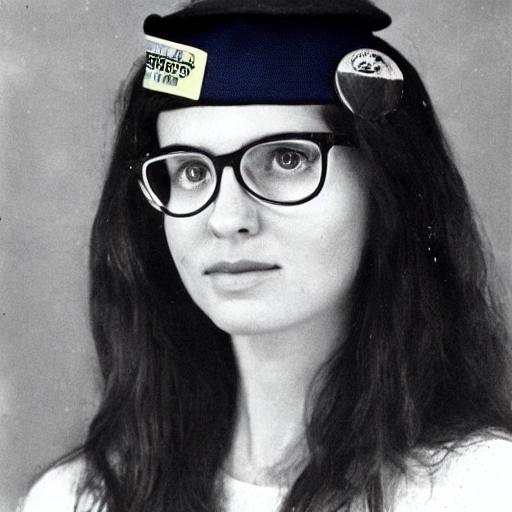}}
  \hspace{0.6cm}
  \subfloat[\scriptsize{A boy wearing glasses \\ A cap}]
  {\includegraphics[width=0.25\textwidth]{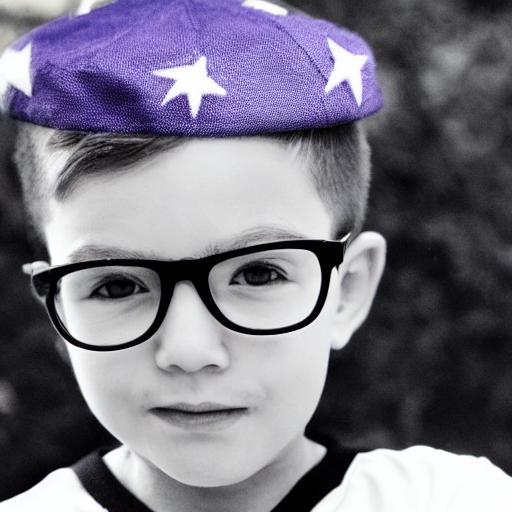}}
  \vspace{0.1cm}
  \subfloat[\scriptsize{A lake \\ A plane}]
  {\includegraphics[width=0.25\textwidth]{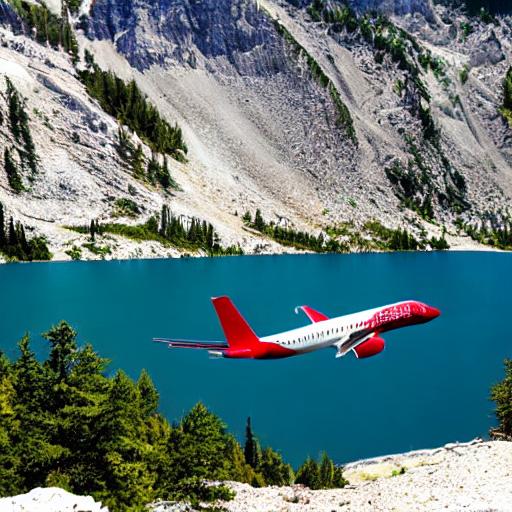}}
  \hspace{0.6cm}
  \subfloat[\scriptsize{A lake \\ A warship}]
  {\includegraphics[width=0.25\textwidth]{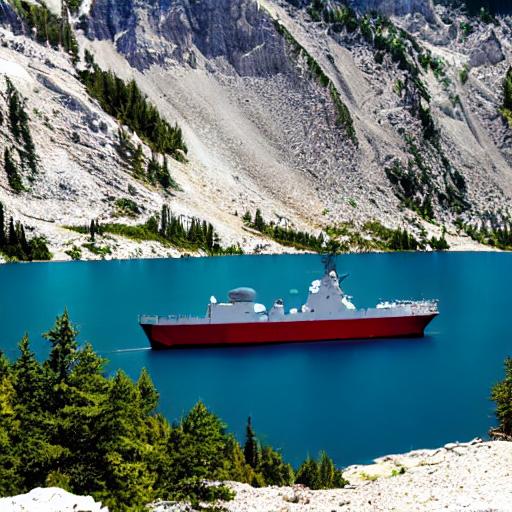}}
  \hspace{0.6cm}
  \subfloat[\scriptsize{A lake \\ A whale}]
  {\includegraphics[width=0.25\textwidth]{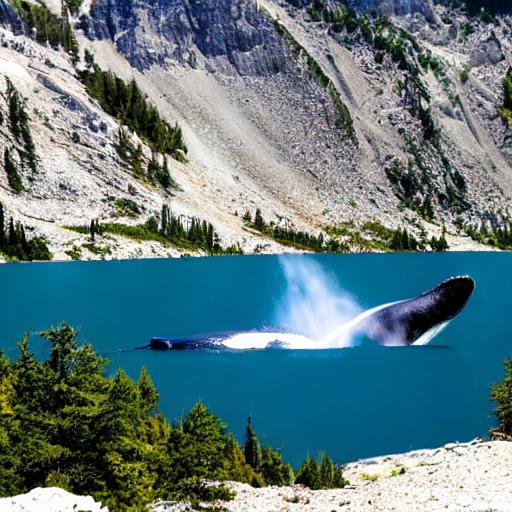}}
  \vspace{0.1cm}
  \subfloat[\scriptsize{A whole round table \\ A basketball}]
  {\includegraphics[width=0.25\textwidth]{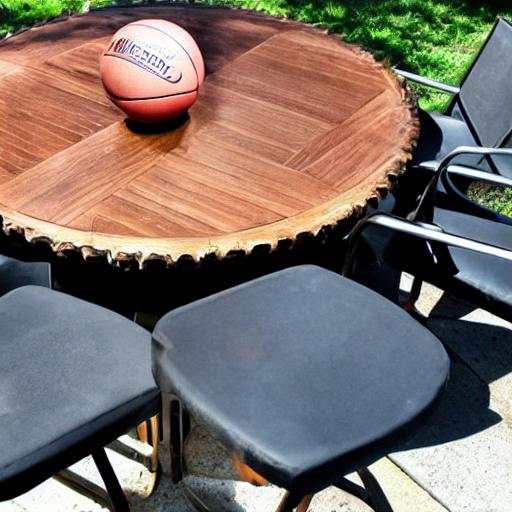}}
  \hspace{0.6cm}
  \subfloat[\scriptsize{A whole round table \\ A pumpkin}]
  {\includegraphics[width=0.25\textwidth]{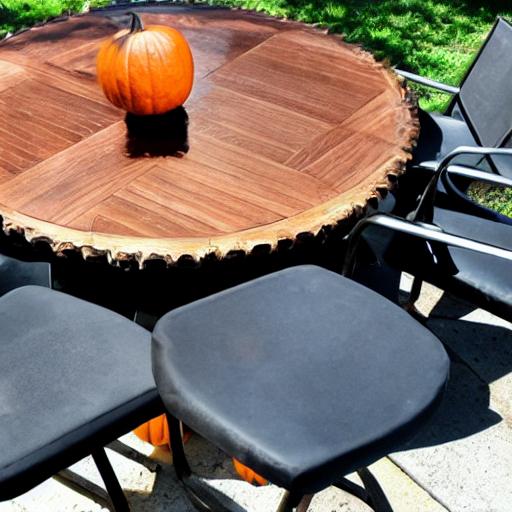}}
  \hspace{0.6cm}
  \subfloat[\scriptsize{A whole round table \\ A orange}]
  {\includegraphics[width=0.25\textwidth]{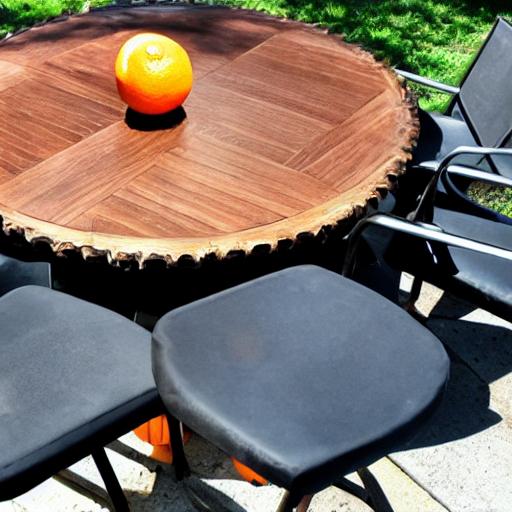}}
\caption{Visualization of ObjectAdd instructed by different prompts $\mathcal{P}$ and words $\mathcal{W}$.}
\label{more_results}
%\vspace{-1.0em}
\end{figure*}

\subsection{Qualitative Comparison}\label{qualita}

As illustrated in Fig.\,\ref{result_compare}, our ObjectAdd method outperforms all other techniques in terms of performance. InstructPix2Pix fails to generate the adding object in most cases, while been unnatural and inconsistent in the case of apples. Both P2P and SD-v1-4 occasionally encounter difficulties in creating the object or yield a low-quality rendition, depending on the ambiguity of the merged prompt. Moreover, they predominantly fail to uphold consistency. As a result, these qualitative results well demonstrate that our ObjectAdd is  superior to the other methods.
Fig.\,\ref{more_results} showcases additional outputs from our ObjectAdd. The first row demonstrates its capability to add various ``hat'' objects to different people, displaying rich details and ensuring a good fit despite the unique head shapes. In the second and third rows, we use the same user-specified mask to add different objects into the same image, showing our method's versatility. In the lake scenarios, we add a plane flying above the water, a warship floating on the surface, and a whale mostly submerged, all interacting with the water in realistic ways, including waves and reflections. These examples show that our ObjectAdd could recognize specific actions and execute accurate interaction even for the same background. The last row shows other examples in other conditions, highlighting and stressing the same advantages of our approach as mentioned.

\begin{figure*}[!t]
  \centering
  \begin{minipage}[t]{0.19\linewidth}
    \centering
    \subfloat[None]{\includegraphics[height=3.2cm]{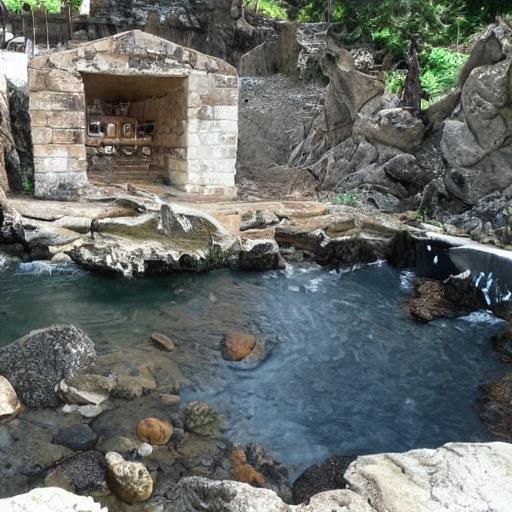}}
    \label{fig:nothing}
  \end{minipage}
  \hfill
  \begin{minipage}[t]{0.19\linewidth}
    \centering
    \subfloat[+ TEC]{\includegraphics[height=3.2cm]{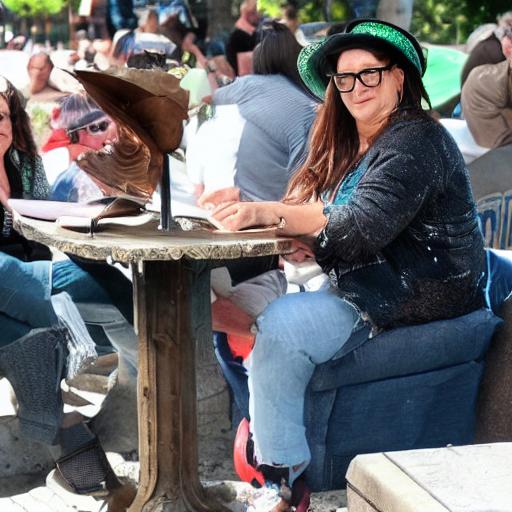}}
    \label{fig:embeddig_merge}
  \end{minipage}
  \hfill
  \begin{minipage}[t]{0.19\linewidth}
    \centering
    \subfloat[+ LII]{\includegraphics[height=3.2cm]{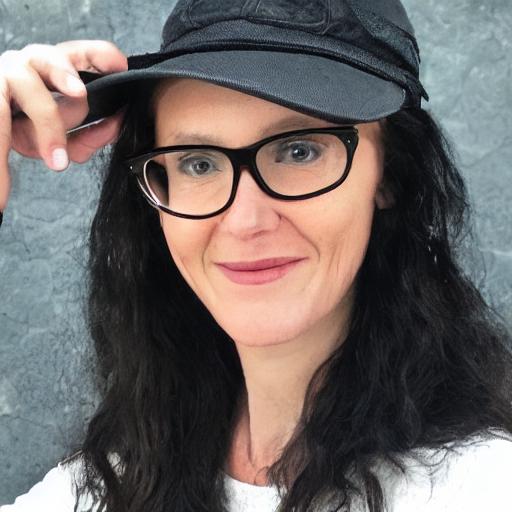}}
    \label{fig:latent_injection}
  \end{minipage}
  \hfill
  \begin{minipage}[t]{0.19\linewidth}
    \centering
    \subfloat[+ AII]{\includegraphics[height=3.2cm]{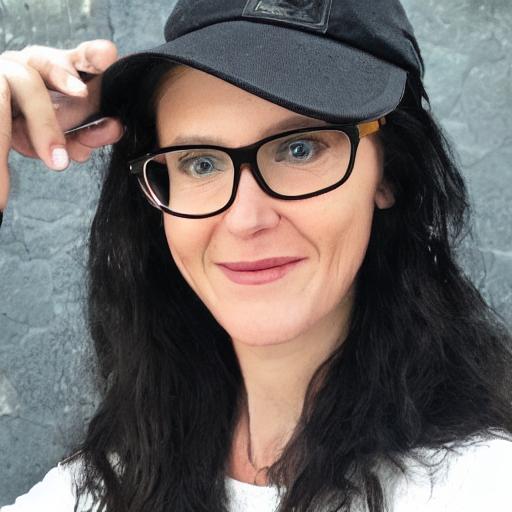}}
    \label{fig:attention_injection}
  \end{minipage}
  \hfill
  \begin{minipage}[t]{0.19\linewidth}
    \centering
    \subfloat[+ PII]{\includegraphics[height=3.2cm]{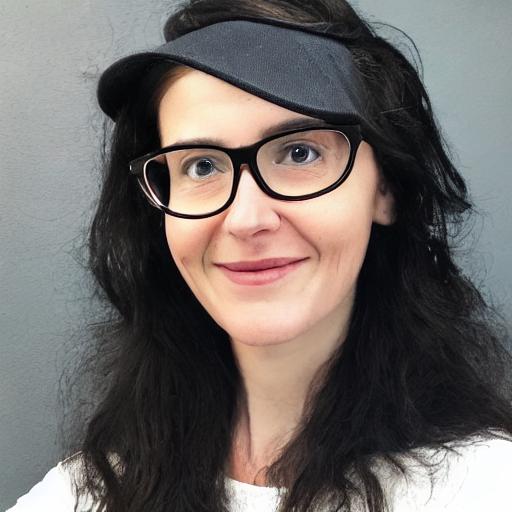}}
    \label{fig:object_replace}
  \end{minipage}
  \caption{Visualizations under ablated settings with $\mathcal{P} = $ ``A woman wearing glasses'' and $\mathcal{W} = $ ``A hat.'' Progressing from (a) to (e), we sequentially incorporate Text Embedding Coalesce (TEC), Latent Information Injection (LII), Attention Information Injection (AII), and Prompted Image Inpainting (PII) techniques.}
  \label{fig:ablation}
  \vspace{-1.0em}
\end{figure*}

\begin{table}[!t]\tabcolsep=0.3cm
\caption{A comparison of quantitative results.}
\centering
 \begin{tabular}{c c c c} 
 \toprule
 & By Pixels ($\downarrow$) & FID ($\downarrow$) & CLIP Score ($\uparrow$) \\ [0.5ex] 
 \midrule
 InstructPix2Pix & 11.393  & 39.559 & 18.129 \\
 \hline
 P2P & 12.833 & 46.106 & 17.809 \\
 \hline
 SD-v1-4 & 49.824 & 157.821 & 18.724 \\
 \hline
 ObjectAdd (Ours) & \textbf{10.219} & 82.437 & \textbf{19.424} \\
 \bottomrule
\end{tabular}
\label{quantitative}
\end{table}

\subsection{Quantitative Comparison}
We compute the three metrics delineated in Sec.\,\ref{setup} for all four methods to compare their quantitative outcomes. The assessment is carried out on the 229 pairs of prompts $\mathcal{P}$ generated images and their corresponding versions with the added word $\mathcal{W}$ described objects, as well as the box that illustrates the required position for object generation. Table\,\ref{quantitative} showcases the test results, wherein our ObjectAdd method surpasses others in general.

First and foremost, our ObjectAdd method demonstrates a remarkable ability to maintain the consistency of non-edited areas, achieving the best performance in ``By Pixels'' compared to other techniques. Additionally, in terms of the ``CLIP Score'' metric, our approach significantly surpasses all other methods. As mentioned, ``By Pixels'' indicates the consistency between the non-masked areas of the edited image and the original one, while the ``CLIP Score'' measures the extent to which the masked area fits the corresponding prompt. Thus, the results of these two metrics suggest that our approach performs the best on both masked and unmasked areas compared to other methods.

The ``FID'' score of our method does not seem satisfactory; however, this is because other methods barely edit the original image. Since FID measures the difference in the distribution between two datasets—original images and edited images—a lower value suggests better consistency between the original and edited images. However, the ``By Pixels'' metric indicates that our approach has the best consistency in the unmasked areas. Therefore, we can conclude that the better FID scores of InstructPix2Pix and P2P compared to our method are due to their limited image editing capabilities, which is consistent with the ``CLIP Score'' metric.

%Secondly, in terms of the ``Hist'' metric, DALL-E 3 and SD-v1-4 trail significantly, while P2P displays a performance closely resembling ours. This can be attributed to the minimal alterations made to the image by P2P, as depicted in Fig.\,\ref{result_compare}. Nonetheless, our proposed ObjectAdd method remains the front-runner. Lastly, when examining the CLIP score, the three comparison methods exhibit similar performance, whereas our ObjectAdd method significantly outperforms them all. This suggests that our approach is more adept at generating the desired object in the appropriate location. These observations are congruent with the visual findings in Fig.\,\ref{result_compare}.

\subsection{Ablation Study}\label{ablation_exp}
We engage in ablation studies to scrutinize the effectiveness of each module introduced in this paper. These include text embedding coalesce, object-driven layout control (incorporating both latent and attention information injection), and prompted image inpainting. We confirm their efficacy by sequentially integrating each module and visually presenting the outputs in Fig.\,\ref{fig:ablation}.
%\footnote{More ablations \emph{w.r.t.} time steps to incorporate object-driven layout control and prompted image inpainting have been provided in the {\color{blue}supplementary material}.}
%

Fig.\,\ref{fig:ablation}(a) clearly demonstrates that the omission of these methodologies leads to outputs that bear no relevance to the prompts. This issue arises due to our approach's reliance on text embedding coalescence for the generation of image embeddings. In the absence of this process, there is no conditioned input for generation, resulting in arbitrary images. As depicted in Fig.\,\ref{fig:ablation}(b), text embedding coalescence facilitates the emergence of principal subjects, such as a woman and a hat. However, without layout control, the output suffers from a lack of consistency. This challenge is largely mitigated by the introduction of latent injection, causing the results to more closely resemble our final output. Yet, the details of the hat appear distorted, exhibiting two brims as shown in Fig.\,\ref{fig:ablation}(c). This issue is rectified through the application of attention injection, which enhances the correlation between the mask area and the object prompt, as evidenced in Fig.\,\ref{fig:ablation}(d). The final obstacle pertains to the background. As illustrated in Fig.\,\ref{fig:ablation}(e), we employ prompted image inpainting to achieve non-edited area consistency.

Further more, we delve deeper into our research by offering additional ablative studies concerning time steps for object-driven layout control (latent information injection and attention information injection), and the specific step for prompted image inpainting. 
Our comprehensive analysis aims to shed more light on the intricate dynamics of these processes and provide a more nuanced understanding of their roles in image generation and manipulation.

\begin{figure*}[!ht]
  \centering
  \begin{minipage}[t]{0.19\linewidth}
    \centering
    \subfloat[\scriptsize{All steps}]
    {\includegraphics[height=3.5cm]{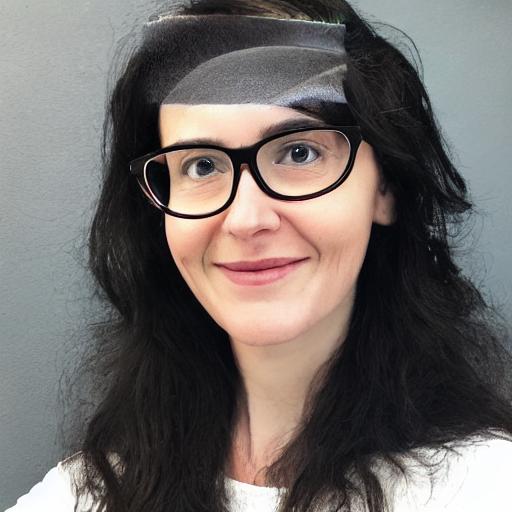}}
  \end{minipage}
  \hfill
  \begin{minipage}[t]{0.19\linewidth}
    \centering
    \subfloat[\scriptsize{First 50$\%$ steps}]
    {\includegraphics[height=3.5cm]{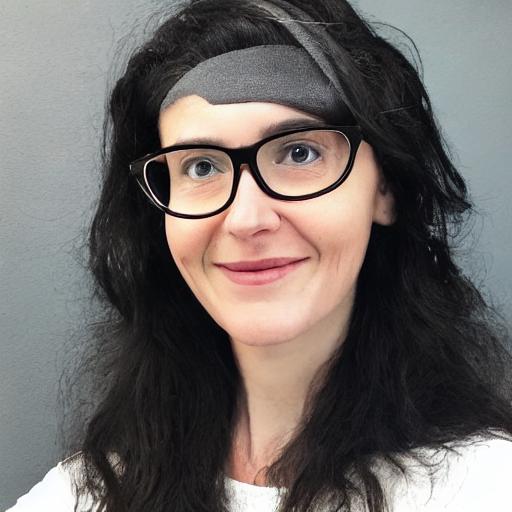}}
  \end{minipage}
  \hfill
  \begin{minipage}[t]{0.19\linewidth}
    \centering
    \subfloat[\scriptsize{First 30$\%$ steps}]
    {\includegraphics[height=3.5cm]{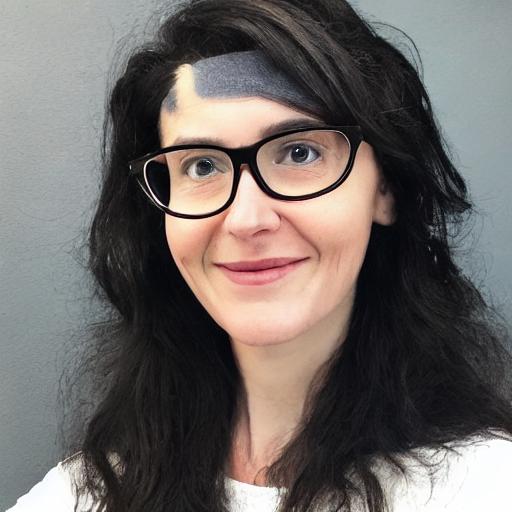}}
  \end{minipage}
  \hfill
  \begin{minipage}[t]{0.19\linewidth}
    \centering
    \subfloat[\scriptsize{First 20$\%$ steps}]
    {\includegraphics[height=3.5cm]{+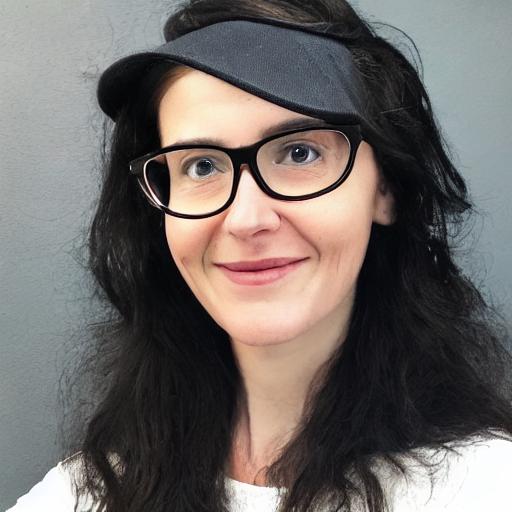}}
  \end{minipage}
  \hfill
  \begin{minipage}[t]{0.19\linewidth}
    \centering
    \subfloat[\scriptsize{First 10$\%$ steps}]
    {\includegraphics[height=3.5cm]{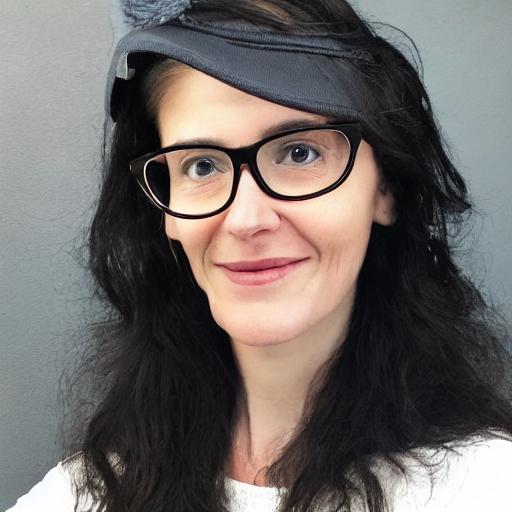}}
  \end{minipage}
  \caption{Visualizations for latent information injection over different time steps under ablated settings with $\mathcal{P} = $ ``A woman wearing glasses'' and $\mathcal{W} = $ ``A hat.'' }
  \label{latent_inj}
  %, only our setting manage to achieve a good performance.
\end{figure*}

\textbf{Latent Information Injection.} In Fig.\,\ref{latent_inj}, we present the outcomes of various latent injection steps, showing the efficacy of this approach in generating high-quality images. As the time steps $t = T \rightarrow 0$, we observe that the optimal performance is achieved when the latent information is injected during the initial 20\% of the time steps. Employing an excessive number of steps for this operation may cause the appearance of unnatural edges and inconsistencies within the object, whereas utilizing too few steps might compromise the overall quality of the generated image.

Through this analysis, it becomes evident that the proper balance of latent information injection is crucial for obtaining visually appealing and accurate results. By carefully adjusting the number of steps, we can strike the perfect balance between image quality and the preservation of essential object features, ultimately leading to a more refined and sophisticated outcome.

\begin{figure*}[!t]
  \centering
  \begin{minipage}[t]{0.19\linewidth}
    \centering
    \subfloat[\scriptsize{All steps}]
    {\includegraphics[height=3.5cm]{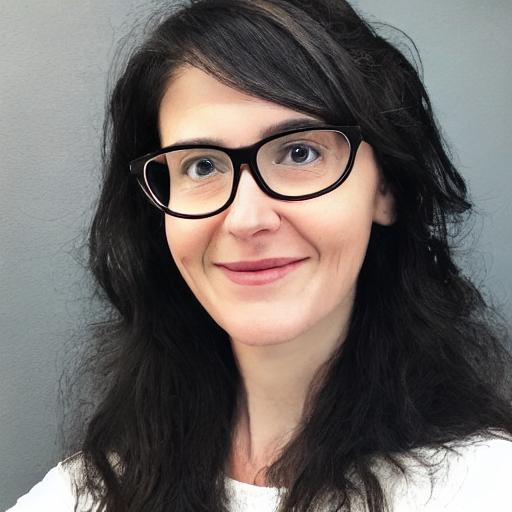}}
  \end{minipage}
  \hfill
  \begin{minipage}[t]{0.19\linewidth}
    \centering
    \subfloat[\scriptsize{First 60$\%$ steps}]
    {\includegraphics[height=3.5cm]{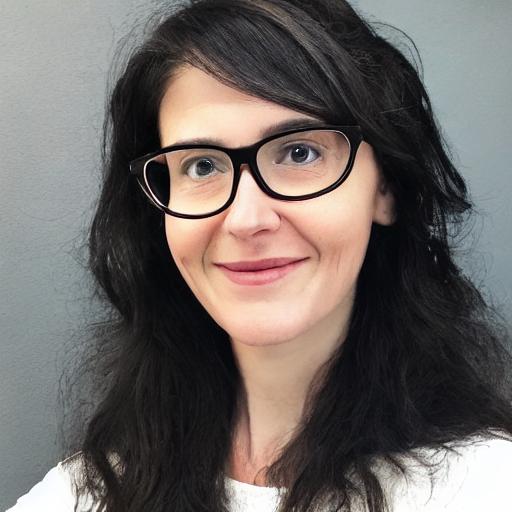}}
  \end{minipage}
  \hfill
  \begin{minipage}[t]{0.19\linewidth}
    \centering
    \subfloat[\scriptsize{First 40$\%$ steps}]
    {\includegraphics[height=3.5cm]{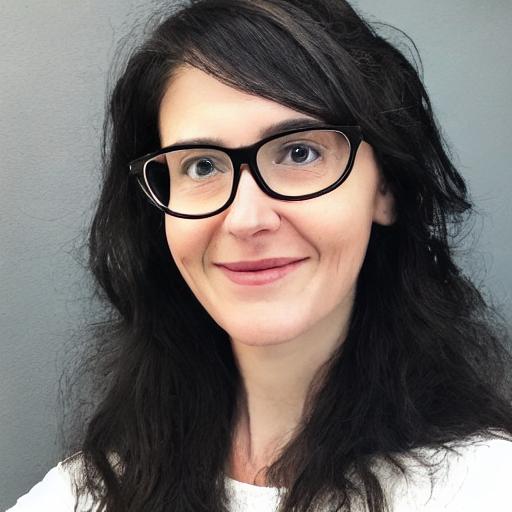}}
  \end{minipage}
  \hfill
  \begin{minipage}[t]{0.19\linewidth}
    \centering
    \subfloat[\scriptsize{First 30$\%$ steps}]
    {\includegraphics[height=3.5cm]{+object_replace_edit_img.jpeg}}
  \end{minipage}
  \hfill
  \begin{minipage}[t]{0.19\linewidth}
    \centering
    \subfloat[\scriptsize{First 20$\%$ steps}]
    {\includegraphics[height=3.5cm]{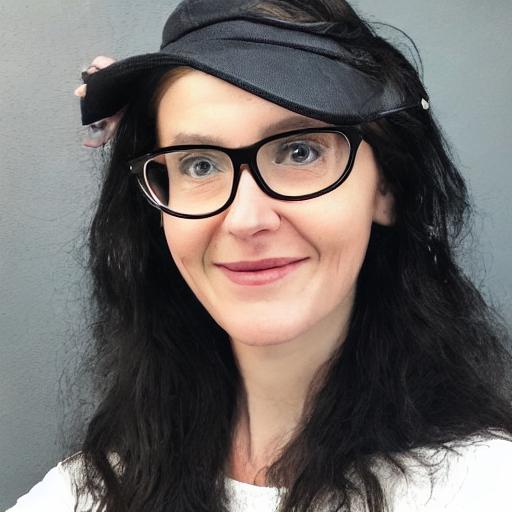}}
  \end{minipage}
  \caption{Visualizations for attention information injection over different time steps under ablated settings with $\mathcal{P} = $ ``A woman wearing glasses'' and $\mathcal{W} = $ ``A hat.'' }
  \label{attention_inj}
  %, it is clear that the time step of attention injection should be exactly as our setting
\end{figure*}

\textbf{Attention Information Injection.} As shown in Fig.\,\ref{attention_inj}, similar to what occurs in latent information injection, too many steps of injection cause the object to disappear, while fewer steps than our setting can lead to poor quality. The most appropriate steps falls into the first 30\%  steps.

%\vspace{-0.5cm}

\begin{figure*}[!hb]
  \centering
  \begin{minipage}[t]{0.19\linewidth}
    \centering
    \subfloat[\scriptsize{Step 30}]
    {\includegraphics[height=3.5cm]{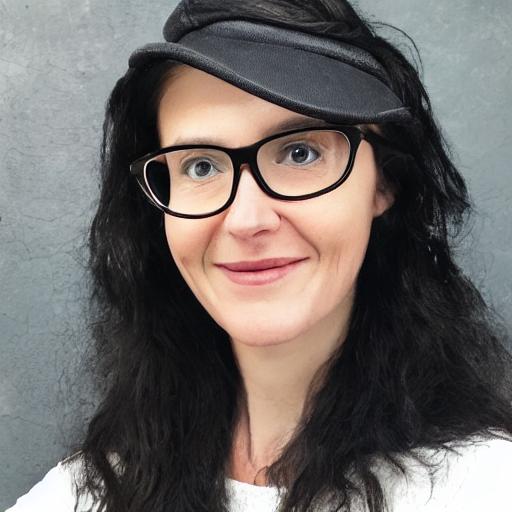}}
  \end{minipage}
  \hfill
  \begin{minipage}[t]{0.19\linewidth}
    \centering
    \subfloat[\scriptsize{Step 25}]
    {\includegraphics[height=3.5cm]{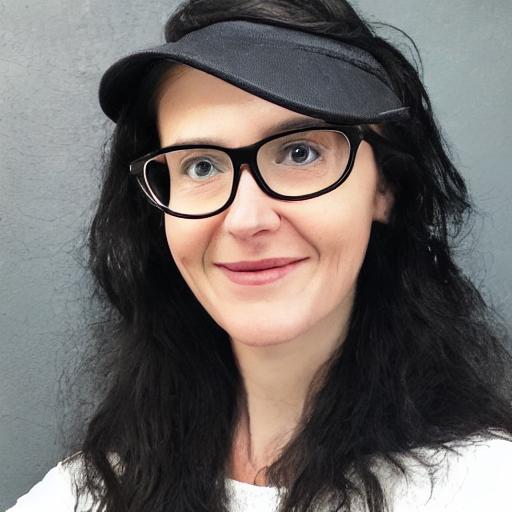}}
  \end{minipage}
  \hfill
  \begin{minipage}[t]{0.19\linewidth}
    \centering
    \subfloat[\scriptsize{Step 20}]
    {\includegraphics[height=3.5cm]{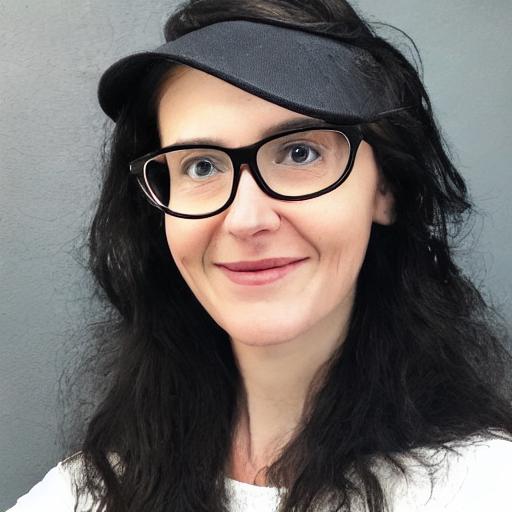}}
  \end{minipage}
  \hfill
  \begin{minipage}[t]{0.19\linewidth}
    \centering
    \subfloat[\scriptsize{Step 15}]
    {\includegraphics[height=3.5cm]{+object_replace_edit_img.jpeg}}
  \end{minipage}
  \hfill
  \begin{minipage}[t]{0.19\linewidth}
    \centering
    \subfloat[\scriptsize{Step 10}]
    {\includegraphics[height=3.5cm]{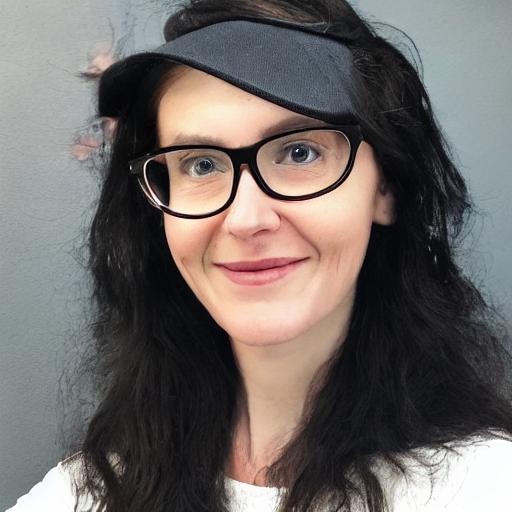}}
  \end{minipage}
  \caption{Visualizations for prompted image inpainting over different time steps under ablated settings with $\mathcal{P} = $ ``A woman wearing glasses'' and $\mathcal{W} = $ ``A hat.'' }
  \label{obj_inj}
  %, image (b) to (d) proved that the time step for this operation could be randomly chose between $0.3 \cdot T$ to $0.5 \cdot T$.
\end{figure*}

\textbf{Prompted Image Inpainting.} In previous sections, we have discussed the optimal selection of the time step $t$ for prompted image inpainting, suggesting that it should fall within the range of $0.3 \cdot T < t < 0.5 \cdot T$. Our experimental findings, as depicted in Fig.\,\ref{obj_inj}, substantiate this claim. Images (b) through (d) exhibit remarkable similarity, indicating that variations in the chosen $t$ values do not significantly impact the final outcomes. Conversely, images (a) and (e) demonstrate inferior performance, signifying that selecting $t$ values beyond the recommended range is ill-advised.

These experimental results corroborate the validity of our hyper-parameter settings and emphasize their importance in achieving optimal results. By adhering to the suggested range for $t$, we can ensure a robust and consistent performance in prompted image inpainting, ultimately leading to superior image quality and enhanced visual fidelity.

\begin{figure*}[ht]
    \centering
    \subfloat[]{\includegraphics[width=0.5\textwidth]{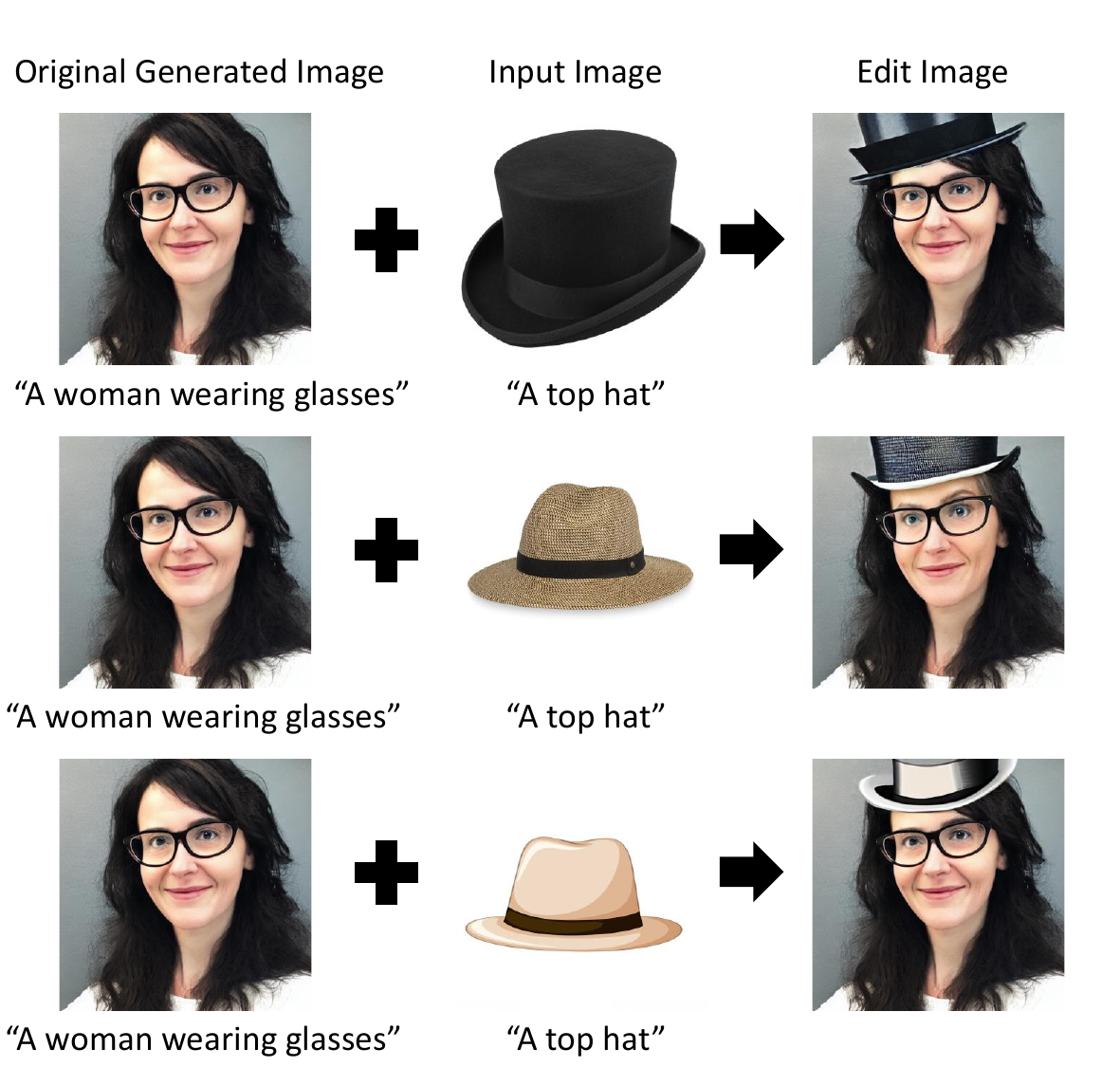}}
    \subfloat[]{\includegraphics[width=0.5\textwidth]{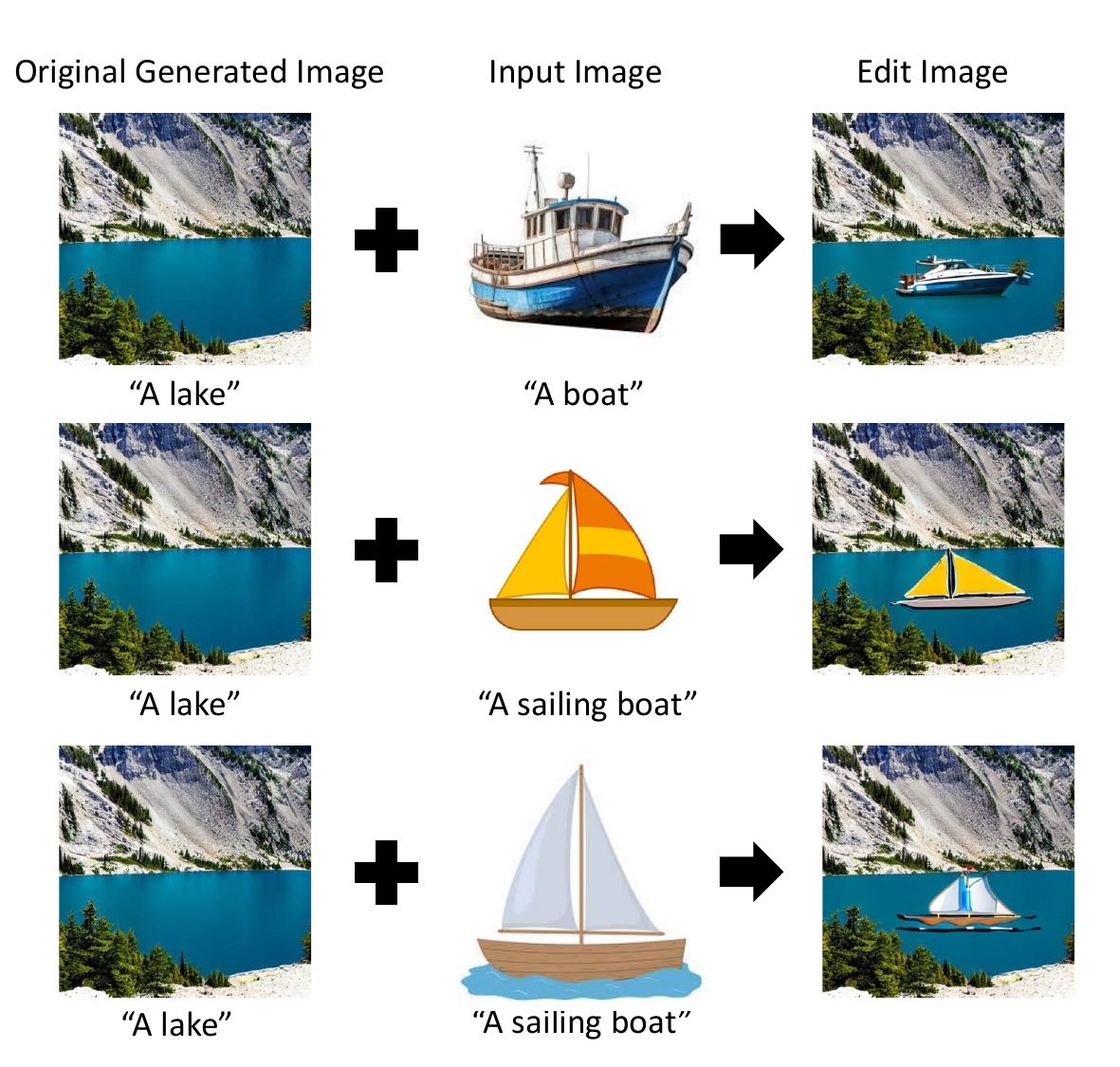}}
    \caption{Example of inserting user specified images.}
    \label{fig:real_imgs_show}
\end{figure*}

\subsection{Performance with a Real Image as Input}\label{real_exp}
As mentioned in previous sections, our approach also allows users to take an image as input for inserting it into the boxing area. In this section, we will display some results of this function and discuss them. As illustrated in Fig.\,\ref{fig:real_imgs_show}, our approach successfully generates the required object in the boxing area while preserving the main features of the input image, such as the color of boats and the texture of top hats. One interesting observation is that the prompt for the inserted image should accurately reflect its content; otherwise, the result may be of poor quality after insertion.

\subsection{Limitation and Future Work}\label{future_works}
While ObjectAdd represents a groundbreaking endeavor, it is not without limitations:
(1) The intricate nature of the approach, such as time steps, prompted image inpainting, and hyperparameter adjustments, constrains its usability for non-experts.
(2) The performance is partially dependent on the pre-trained SD-v1-4 model, which may compromise its effectiveness in complex scenarios, such as ``a dog standing upon a man's head.''
(3) The evaluation is based on a limited input set, potentially not fully showcasing the method's capabilities.

Moving forward, the focus will be on developing a refined approach that ensures seamless accessibility for non-specialists. Besides, exploring advanced open models like SD-v3 version~\cite{esser2024scaling} is crucial. Finally, creating a suite of automated evaluation tools for comprehensive assessment is of paramount importance. Besides, it would be an interesting idea to transform input image to prompt automatically using a pre-trained model like BLIP~\cite{li2022blip}, which might make our method more practical for using.

\section{Conclusion}
In this study, we have presented a groundbreaking training-free methodology for incorporating objects into a user-defined area of a diffusion image while preserving the remainder of the image with minimal alterations. As a pioneering effort in this domain, our proposed ObjectAdd technique employs embedding-level concatenation to guarantee accurate text embedding coalescence, object-driven layout control through latent and attention injection to ensure objects are placed within the user-specified area, and prompted image inpainting utilizing attention refocusing and object expansion strategies to maintain the integrity of the rest of the image. A thorough combination of qualitative and quantitative comparisons with closely related works, such as DALL-E 3, P2P, InstructPix2Pix, and SD-v1-4, have convincingly showcased the efficacy of our innovative ObjectAdd methodology.

\section*{Acknowledgments}
This work was supported by National Science and Technology Major Project (No. 2022ZD0118202), the National Science Fund for Distinguished Young Scholars (No.62025603), the National Natural Science Foundation of China (No. U21B2037, No. U22B2051, No. 62176222, No. 62176223, No. 62176226, No. 62072386, No. 62072387, No. 62072389, No. 62002305 and No. 62272401), and the Natural Science Foundation of Fujian Province of China (No.2021J01002,  No.2022J06001).

\bibliographystyle{IEEEtran}
\bibliography{egbib}

@inproceedings{pan2023effective,
  title={Effective Real Image Editing with Accelerated Iterative Diffusion Inversion},
  author={Pan, Zhihong and Gherardi, Riccardo and Xie, Xiufeng and Huang, Stephen},
  booktitle={Proceedings of the IEEE/CVF Conference on Computer Vision and Pattern Recognition},
  year={2023}
}

@inproceedings{couairon2023diffedit,
  title={DiffEdit: Diffusion-based Semantic Image Editing with Mask Guidance},
  author={Couairon, Guillaume and Verbeek, Jakob and Schwenk, Holger and Cord, Matthieu},
  booktitle={The Twelfth International Conference on Learning Representations},
  year={2023}
}

@inproceedings{radford2021learning,
  title={Learning transferable visual models from natural language supervision},
  author={Radford, Alec and Kim, Jong Wook and Hallacy, Chris and Ramesh, Aditya and Goh, Gabriel and Agarwal, Sandhini and Sastry, Girish and Askell, Amanda and Mishkin, Pamela and Clark, Jack and others},
  booktitle={International Conference on Machine Learning},
  pages={8748--8763},
  year={2021}
}

@inproceedings{rombach2022high,
  title={High-resolution image synthesis with latent diffusion models},
  author={Rombach, Robin and Blattmann, Andreas and Lorenz, Dominik and Esser, Patrick and Ommer, Bj{\"o}rn},
  booktitle={Proceedings of the IEEE/CVF Conference on Computer Vision and Pattern Recognition},
  pages={10684--10695},
  year={2022}
}

@inproceedings{hertz2022prompt,
  title={Prompt-to-Prompt Image Editing with Cross-Attention Control},
  author={Hertz, Amir and Mokady, Ron and Tenenbaum, Jay and Aberman, Kfir and Pritch, Yael and Cohen-or, Daniel},
  booktitle={The Eleventh International Conference on Learning Representations},
  year={2022}
}

@inproceedings{chen2023trainingfree,
  title={Training-free layout control with cross-attention guidance},
  author={Chen, Minghao and Laina, Iro and Vedaldi, Andrea},
  booktitle={Proceedings of the IEEE/CVF Winter Conference on Applications of Computer Vision},
  pages={5343--5353},
  year={2024}
}

@inproceedings{zhai2024investigating,
  title={Investigating the Catastrophic Forgetting in Multimodal Large Language Model Fine-Tuning},
  author={Zhai, Yuexiang and Tong, Shengbang and Li, Xiao and Cai, Mu and Qu, Qing and Lee, Yong Jae and Ma, Yi},
  booktitle={Conference on Parsimony and Learning},
  pages={202--227},
  year={2024}
}

@inproceedings{densediffusion,
  title={Dense text-to-image generation with attention modulation},
  author={Kim, Yunji and Lee, Jiyoung and Kim, Jin-Hwa and Ha, Jung-Woo and Zhu, Jun-Yan},
  booktitle={Proceedings of the IEEE/CVF International Conference on Computer Vision},
  pages={7701--7711},
  year={2023}
}

@inproceedings{patashnik2023localizing,
    author    = {Patashnik, Or and Garibi, Daniel and Azuri, Idan and Averbuch-Elor, Hadar and Cohen-Or, Daniel},
    title     = {Localizing Object-level Shape Variations with Text-to-Image Diffusion Models},
    booktitle = {Proceedings of the IEEE/CVF International Conference on Computer Vision},
    year      = {2023}
}

@inproceedings{xie2023boxdiff,
  title={Boxdiff: Text-to-image synthesis with training-free box-constrained diffusion},
  author={Xie, Jinheng and Li, Yuexiang and Huang, Yawen and Liu, Haozhe and Zhang, Wentian and Zheng, Yefeng and Shou, Mike Zheng},
  booktitle={Proceedings of the IEEE/CVF International Conference on Computer Vision},
  pages={7452--7461},
  year={2023}
}

@inproceedings{zhang2023iti,
  title={ITI-GEN: Inclusive Text-to-Image Generation},
  author={Zhang, Cheng and Chen, Xuanbai and Chai, Siqi and Wu, Chen Henry and Lagun, Dmitry and Beeler, Thabo and De la Torre, Fernando},
  booktitle={Proceedings of the IEEE/CVF International Conference on Computer Vision},
  pages={3969--3980},
  year={2023}
}

@inproceedings{li2023gligen,
  title={Gligen: Open-set grounded text-to-image generation},
  author={Li, Yuheng and Liu, Haotian and Wu, Qingyang and Mu, Fangzhou and Yang, Jianwei and Gao, Jianfeng and Li, Chunyuan and Lee, Yong Jae},
  booktitle={Proceedings of the IEEE/CVF Conference on Computer Vision and Pattern Recognition},
  pages={22511--22521},
  year={2023}
}

@inproceedings{feng2022training,
  title={Training-Free Structured Diffusion Guidance for Compositional Text-to-Image Synthesis},
  author={Feng, Weixi and He, Xuehai and Fu, Tsu-Jui and Jampani, Varun and Akula, Arjun Reddy and Narayana, Pradyumna and Basu, Sugato and Wang, Xin Eric and Wang, William Yang},
  booktitle={The Eleventh International Conference on Learning Representations},
  year={2022}
}

@inproceedings{zhang2023adding,
  title={Adding conditional control to text-to-image diffusion models},
  author={Zhang, Lvmin and Rao, Anyi and Agrawala, Maneesh},
  booktitle={Proceedings of the IEEE/CVF International Conference on Computer Vision},
  pages={3836--3847},
  year={2023}
}

@inproceedings{cao_2023_masactrl,
    title     = {MasaCtrl: Tuning-Free Mutual Self-Attention Control for Consistent Image Synthesis and Editing},
    author    = {Cao, Mingdeng and Wang, Xintao and Qi, Zhongang and Shan, Ying and Qie, Xiaohu and Zheng, Yinqiang},
    booktitle = {Proceedings of the IEEE/CVF International Conference on Computer Vision},
    pages     = {22560--22570},
    year      = {2023}
}

@inproceedings{heusel2017gans,
  title={GANs trained by a two time-scale update rule converge to a local Nash equilibrium},
  author={Heusel, Martin and Ramsauer, Hubert and Unterthiner, Thomas and Nessler, Bernhard and Hochreiter, Sepp},
  booktitle={Advances in Neural Information Processing Systems},
  pages={6626--6637},
  year={2017}
}

@inproceedings{li2022blip,
  title={Blip: Bootstrapping language-image pre-training for unified vision-language understanding and generation},
  author={Li, Junnan and Li, Dongxu and Xiong, Caiming and Hoi, Steven},
  booktitle={International conference on machine learning},
  pages={12888--12900},
  year={2022}
}

@inproceedings{brooks2023instructpix2pix,
  title={Instructpix2pix: Learning to follow image editing instructions},
  author={Brooks, Tim and Holynski, Aleksander and Efros, Alexei A},
  booktitle={Proceedings of the IEEE/CVF Conference on Computer Vision and Pattern Recognition},
  pages={18392--18402},
  year={2023}
}

@book{soille1999morphological,
  title={Morphological image analysis: principles and applications},
  author={Soille, Pierre and others},
  volume={2},
  year={1999},
  publisher={Springer}
}

@misc{taited2023CLIPScore,
  author={SUN Zhengwentai},
  year={2023},
  title={{clip-score: CLIP Score for PyTorch}},
  note={\url{https://github.com/taited/clip-score}},
}

@misc{midjourney,
    author= {{Midjourney}},
    year  = {2024},
    title = {Midjourney},
    note  = {\url{https://www.midjourney.com/home}, 
             Last accessed on 2024-2-27},
}

@misc{chatGPT,
    author= {{Openai}},
    year  = {2024},
    title = {ChatGPT},
    note  = {\url{https://chat.openai.com/}, 
             Last accessed on 2024-2-27},
}

@misc{photoshop,
    author= {{Adobe}},
    year  = {2024},
    title = {Photoshop},
    note  = {\url{https://www.adobe.com/products/photoshop.html}, 
             Last accessed on 2024-2-27},
}

@article{garibi2024renoise,
  title={ReNoise: Real Image Inversion Through Iterative Noising},
  author={Garibi, Daniel and Patashnik, Or and Voynov, Andrey and Averbuch-Elor, Hadar and Cohen-Or, Daniel},
  journal={arXiv preprint arXiv:2403.14602},
  year={2024}
}

@article{ho2020denoising,
  title={Denoising diffusion probabilistic models},
  author={Ho, Jonathan and Jain, Ajay and Abbeel, Pieter},
  journal={Advances in Neural Information Processing Systems},
  pages={6840--6851},
  year={2020}
}

@article{betker2023improving,
  title={Improving image generation with better captions},
  author={Betker, James and Goh, Gabriel and Jing, Li and Brooks, Tim and Wang, Jianfeng and Li, Linjie and Ouyang, Long and Zhuang, Juntang and Lee, Joyce and Guo, Yufei and others},
  journal={Computer Science. https://cdn. openai. com/papers/dall-e-3. pdf},
  pages={3},
  year={2023}
}

@article{chefer2023attend,
  title={Attend-and-excite: Attention-based semantic guidance for text-to-image diffusion models},
  author={Chefer, Hila and Alaluf, Yuval and Vinker, Yael and Wolf, Lior and Cohen-Or, Daniel},
  journal={ACM Transactions on Graphics},
  pages={1--10},
  year={2023},
}

@article{shi2023dragdiffusion,
    title={DragDiffusion: Harnessing Diffusion Models for Interactive Point-based Image Editing},
    author={Shi, Yujun and Xue, Chuhui and Pan, Jiachun and Zhang, Wenqing and Tan, Vincent YF and Bai, Song},
    journal={ArXiv Preprint arXiv:2306.14435},
    year={2023}
}

@article{esser2024scaling,
    title={Scaling Rectified Flow Transformers for High-Resolution Image Synthesis}, 
    author={Patrick Esser and Sumith Kulal and Andreas Blattmann and Rahim Entezari and Jonas Müller and Harry Saini and Yam Levi and Dominik Lorenz and Axel Sauer and Frederic Boesel and Dustin Podell and Tim Dockhorn and Zion English and Kyle Lacey and Alex Goodwin and Yannik Marek and Robin Rombach},
    journal={ArXiv Preprint arXiv:2403.03206},
    year={2024}
}

@article{vaswani2017attention,
  title={Attention is all you need},
  author={Vaswani, Ashish and Shazeer, Noam and Parmar, Niki and Uszkoreit, Jakob and Jones, Llion and Gomez, Aidan N and Kaiser, {\L}ukasz and Polosukhin, Illia},
  journal={Advances in Neural Information Processing Systems},
  year={2017}
}

\end{document}